\documentclass[conference]{IEEEtran}
\usepackage{times}

\usepackage[numbers]{natbib}
\usepackage{multicol}
\usepackage[bookmarks=true]{hyperref}

\usepackage{graphics} % for pdf, bitmapped graphics files
\usepackage{epsfig} % for postscript graphics files
\usepackage{times} % assumes new font selection scheme installed
\usepackage{amsmath} % assumes amsmath package installed
\usepackage{amssymb}  % assumes amsmath package installed
\usepackage[export]{adjustbox}
\usepackage{subfig}
\usepackage{graphicx}
\usepackage{wrapfig}
\usepackage{xcolor}
\usepackage{hyperref}
\usepackage{mathtools}
\usepackage{natbib}
\usepackage{tabu, booktabs}
\usepackage{tabu, booktabs}

\usepackage{url}

\usepackage[numbers]{natbib}
\usepackage{balance}

\newsavebox{\measurebox}

\hypersetup{pdftitle={Differentiable Particle Filters: End-to-End Learning with Algorithmic Priors},pdfauthor={Rico Jonschkowski and Divyam Rastogi and Oliver Brock}}

\begin{document}

%\title{Network Architectures from Algorithms: Learning State Estimation with Differentiable Particle Filters}
%\title{{\bf Learning for Algorithms}: Optimizing Models End-to-End with Differentiable Particle Filters}
%\title{{\bf Network Structures from Algorithms}: End-to-End Learning with Differentiable Particle Filters}
%\title{{ Differentiable Particle Filters}: Learning of State Estimation End-to-End with Algorithmic Priors}
%\title{{\bf Differentiable Particle Filters}:\\End-to-End Learning in a Differentiable Algorithm}
\title{{\bf Differentiable Particle Filters}:\\End-to-End Learning with Algorithmic Priors}

% You will get a Paper-ID when submitting a pdf file to the conference system
%\author{Author Names Omitted for Anonymous Review. Paper-ID [add your ID here]}

%\author{\authorblockN{Rico Jonschkowski}
%\authorblockA{School of Electrical and\\Computer Engineering\\
%Georgia Institute of Technology\\
%Atlanta, Georgia 30332--0250\\
%Email: mshell@ece.gatech.edu}
%\and
%\authorblockA{Twentieth Century Fox\\
%\and
%\authorblockN{James Kirk\\ and Montgomery Scott}
%\authorblockA{Starfleet Academy\\
%San Francisco, California 96678-2391\\

\author{\authorblockN{Rico Jonschkowski, Divyam Rastogi, and Oliver Brock}
\authorblockA{Robotics and Biology Laboratory, Technische Universit\"at Berlin, Germany}}

% avoiding spaces at the end of the author lines is not a problem with
% conference papers because we don't use \thanks or \IEEEmembership

%Homer Simpson\authorrefmark{2},
%Montgomery Scott\authorrefmark{3} and
%Eldon Tyrell\authorrefmark{4}}
%Telephone: (800) 555--1212, Fax: (888) 555--1212}
%\authorblockA{\authorrefmark{4}Tyrell Inc., 123 Replicant Street, Los Angeles, California 90210--4321}}

\maketitle

\begin{abstract}
We present differentiable particle filters (DPFs): a differentiable implementation of the particle filter algorithm with learnable motion and measurement models. Since DPFs are end-to-end differentiable, we can efficiently train their models by optimizing end-to-end state estimation performance, rather than proxy objectives such as model accuracy. DPFs encode the structure of recursive state estimation with prediction and measurement update that operate on a probability distribution over states. This structure represents an algorithmic prior that improves  learning performance in state estimation problems while enabling explainability of the learned model. Our experiments on simulated and real data show substantial benefits from end-to-end learning with algorithmic priors, e.g.~reducing error rates by $\sim$80\%. Our experiments also show that, unlike long short-term memory networks, DPFs learn localization in a policy-agnostic way and thus greatly improve generalization. Source code is available at \url{https://github.com/tu-rbo/differentiable-particle-filters}.
\end{abstract}

%Problem-specific algorithms and generic machine learning approaches have complementary strengths and weaknesses, trading-off data efficiency and generality. To find the right balance between these, we propose to use problem-specific information encoded in algorithms together with the ability to learn details about the problem-instance from data. We demonstrate this approach in the context of state estimation in robotics, where 

\IEEEpeerreviewmaketitle

%===============================================================================

%\todo{replace 8 lines with abstract} The Bayes' filter algorithm captures the structure of state estimation in robotics. To apply Bayes' filters to specific problems, we need to pick an approximate implementation (e.g.~particle filters) and we need to define or learn a motion model and a measurement model. Hypothesis: The mathematically optimal models that describe measurement and motion are not the best models for state estimation performance. By making the particle filter differentiable, we can optimize the models for end-to-end performance using backpropagation. Compared to generic differentiable functions such as LSTMs, the Bayes' filter structure improves generalization.

%\keywords{End-to-end learning} 

%===============================================================================
\section{Introduction}
\label{sec:introduction}
%===============================================================================

End-to-end learning tunes all parts of a learnable system for end-to-end performance---which is what we ultimately care about---instead of optimizing each part individually. End-to-end learning excels when the right objectives for individual parts are not known; it therefore has significant potential in the context of complex robotic systems.

Compared to learning each part of a system individually, end-to-end learning puts fewer constraints on the individual parts, which can improve performance but can also lead to overfitting. We must therefore balance end-to-end learning with regularization by incorporating appropriate priors. Priors can be encoded in the form of differentiable network architectures. By defining the network architecture and its learnable parameters, we restrict the hypothesis space and thus regularize learning. At the same time, the differentiability of the network allows all of its parts to adapt to each other and to optimize their parameters for end-to-end performance.

This approach has been very successful in computer vision. Highly engineered vision pipelines are outperformed by convolutional networks trained end-to-end~\citep{he_deep_2015}. But it only works because convolutional networks~\citep{lecun_backpropagation_1989} encode priors in the network architecture that are suitable for computer vision---a hierarchy of local filters shared across the image. Problems in robotics possess additional structure, for example in physical interactions with the environment. Only by exploiting all available structure will we be able to realize the full potential of end-to-end learning in robotics.

\emph{But how can we find more architectures like the convolutional network for robotics?} Roboticists have captured problem structure in the form of algorithms, often combined with models of the specific task. By making these algorithms differentiable and their models learnable, we can turn robotic algorithms into network architectures. This approach enables end-to-end learning while also encoding prior knowledge from algorithms, which we call \emph{algorithmic priors}.

Here, we apply \emph{end-to-end learning with algorithmic priors} to state estimation in robotics. In this problem, a robot needs to infer the latent state from its observations and actions. Since a single observation can be insufficient to estimate the state, the robot needs to integrate uncertain information over time.

\begin{figure}[t]
\centering
\includegraphics[width=0.7\columnwidth]{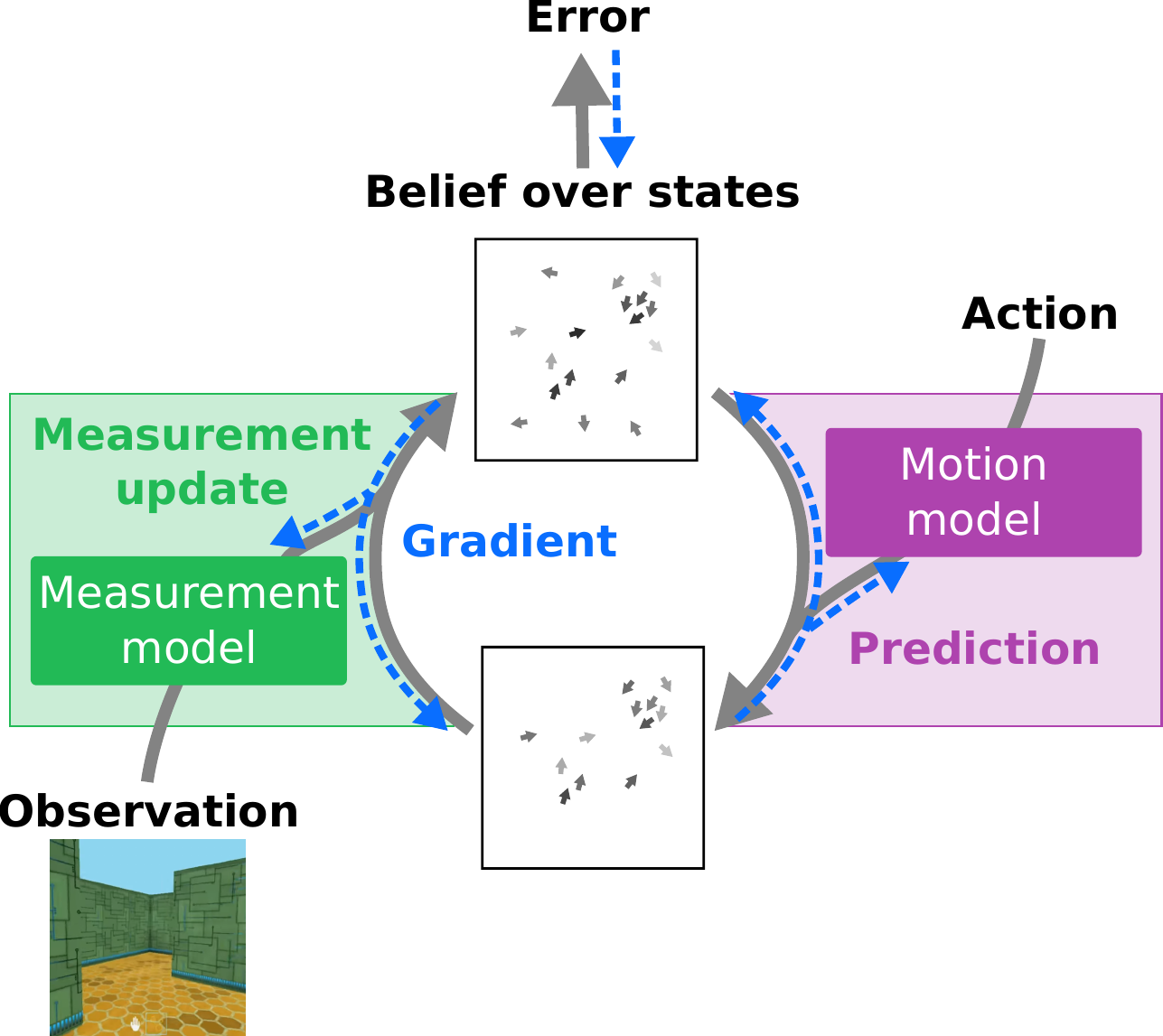} 
\caption{{\bf Differentiable particle filters.} Models can be learned end-to-end by backpropagation through the algorithm.}
\label{fig:overview_simple}
\end{figure}

Given the standard assumptions for this problem, \emph{Bayes filters} provide the provably optimal algorithmic structure for solving it~\citep{thrun_probabilistic_2005}, recursively updating a probability distribution over states with prediction and measurement update using task-specific motion and measurement models. The \emph{differentiable particle filter} (DPF) is an end-to-end differentiable implementation of the particle filter---a Bayes filter that represents probability distributions with samples---with learnable motion and measurement models (see Fig.~\ref{fig:overview_simple}).

Since DPFs are differentiable, we can learn their models end-to-end to optimize state estimation performance. Our experiments show that end-to-end learning improves performance compared to using models optimized for accuracy. Interestingly, end-to-end learning in DPFs re-discovers what roboticists found out via trial and error: that overestimating uncertainty is beneficial for filtering performance \cite[p.~118]{thrun_probabilistic_2005}.

%The algorithmic prior encoded in DPFs has multiple advantages for state estimation. 

Since DPFs use the Bayes filter algorithm as a prior, they have a number of advantages. First, even with end-to-end learning, DPFs remain explainable---we can examine the learned models and their interaction. Second, the algorithmic prior regularizes learning, which greatly improves performance in state estimation. Compared to generic long short-term memory networks (LSTMs) \citep{hochreiter_long_1997}, DPFs reduce the error rate by $\sim$80\% or require 87\% less training data for the same error rate. And finally, the algorithmic prior improves generalization: while LSTMs fail when tested with a different policy than used for training, DPFs are robust to changing the policy.

%We evaluate DPFs for localization in simulated 3D environments from raw visual input. 

%Our results show that end-to-end learning improves performance compared to learning models individually, while the combination of both learning schemes is even more powerful. DPFs also re-discover what roboticists found out via trial and error: that over-estimating uncertainty is beneficial for filtering performance. 

%Compared to generic long short-term memory networks \citep[LSTM,][]{hochreiter_long_1997}, DPFs show substantial advantages, reducing the error rate by $\sim$80\% or requiring only 13\% of the training data. And while LSTMs fail when tested with a different policy than used for training, DPFs are robust to changing the robot's policy.

%learn localization that only works with the training policy, DPFs are robust to changing the .
%While LSTMs learn a localization strategy that only works for the training policy, DPFs learn localization in a way that is agnostic to the robot's policy.

%===============================================================================
\section{Related Work}
\label{sec:related_work}
%===============================================================================

There is a surge of recent work that combines algorithmic priors and end-to-end learning for planning and state estimation with histogram-based and Gaussian belief representations.

% Planning under FULL OBSERVABILITY

\paragraph*{Planning with known state}
Tamar et al. \citep{tamar_value_2016} introduced value iteration networks, a differentiable planning algorithm with models that can be optimized for value iteration. Their key insight is that value iteration in a grid based state space can be represented by convolutional neural networks. Silver et al. \citep{silver_predictron:_2017} proposed the predictron, a differentiable embedding of the TD($\lambda$) algorithm in a learned state space. Okada and Aoshima \citep{okada_path_2017} proposed path integral networks, which encode an optimal control algorithm to learn continuous tasks.

\paragraph*{State estimation (and planning) with histograms}
Jonschkowski and Brock \citep{Jonschkowski-16-NIPS-WS} introduced the end-to-end learnable histogram filter, a differentiable Bayes filter that represents the belief with a histogram. Shankar et al. \citep{shankar_reinforcement_2016} and Karkus et al. \cite{karkus_qmdp-net:_2017} combined histogram filters and QMDP planners in a differentiable network for planning in partially observable environments. Gupta et al. \citep{gupta_cognitive_2017} combined differentiable mapping and planning in a network architecture for navigation in novel environments. All of these approaches use convolution to operate on a grid based state space.

\paragraph*{State estimation with Gaussians}
Harnooja et al. \citep{haarnoja_backprop_2016} presented a differentiable Kalman filter with a Gaussian belief and an end-to-end learnable measurement model from visual input. Watter et al. \citep{watter_embed_2015} and Karl et al. \citep{karl_deep_2017} learn a latent state space that facilitates prediction. These approaches use (locally) linear dynamics models and  Gaussian beliefs.

Related work has established how to operate on \emph{histogram-based} belief representations using convolution and how to work with \emph{Gaussian} beliefs using linear operations. We build on this work and extend its scope to include \emph{sample-based} algorithms, such as particle filters. Sample-based representations can be advantageous because they can represent multi-modal distributions (unlike Gaussians) while focusing the computational effort on states of high probability (unlike histograms). But sample-based representations introduce new challenges for differentiable implementations, e.g. generating samples from networks, performing density estimation to compute gradients, and handling non-differentiable resampling. These are the challenges that we tackle in this paper.

%======================================================================
\section{Background: Bayes Filters and Their Particle-Based Approximation}
%======================================================================

We consider the problem of estimating a latent \emph{state} $\boldsymbol{s}$ from a history of \emph{observations} $\boldsymbol o$ and \emph{actions} $\boldsymbol a$, e.g. a robot's pose from camera images and odometry. To handle uncertainty, we estimate a probability distribution over the current state $\boldsymbol{s}_t$ conditioned on the history of observations $\boldsymbol o_{1:t}$ and actions $\boldsymbol a_{1:t}$, which is called \emph{belief}, $\text{bel}(\boldsymbol{s}_t) = p( \boldsymbol s_t|\boldsymbol a_{1:t}, \boldsymbol o_{1:t}).$

%-------------------------------------------------------------------------------
\subsection{Bayes Filters}
%-------------------------------------------------------------------------------

\begin{figure}[t]
\centering
\includegraphics[width=0.6\columnwidth]{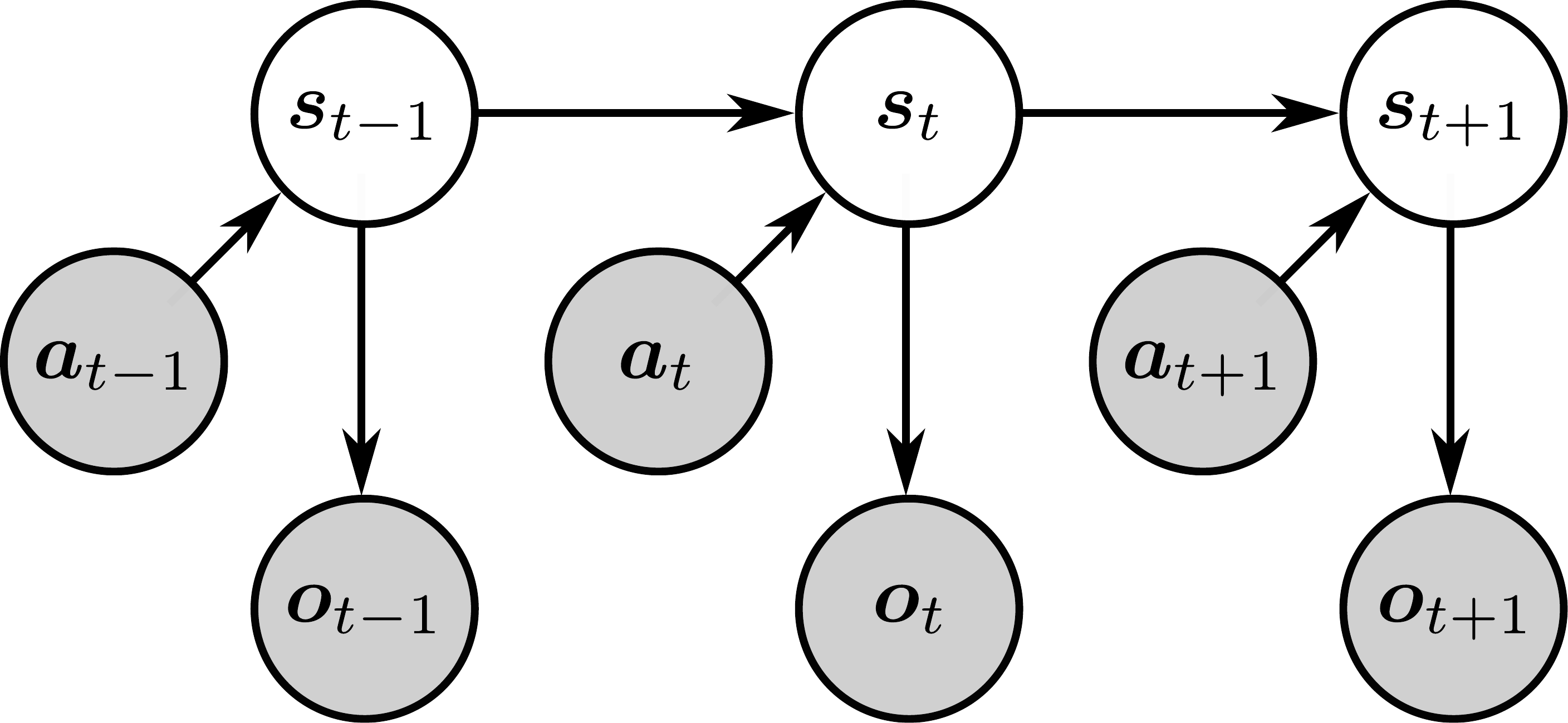}
\caption{\bf Graphical model for state estimation}
\label{fig:graphical_model}
\end{figure}

If we assume that our problem factorizes as shown in Fig.~\ref{fig:graphical_model}, the \emph{Bayes filter} algorithm solves it optimally \citep{thrun_probabilistic_2005} by making use of the Markov property of the state and the conditional independence of observations and actions. From the Markov property follows that the last belief $\text{bel}(\boldsymbol s_{t-1})$ summarizes all information contained in the history of observations $\boldsymbol o_{1:t-1}$ and actions $\boldsymbol a_{1:t-1}$ that is relevant for predicting the future. Accordingly, the Bayes filter computes $\text{bel}(\boldsymbol s_t)$ recursively from $\text{bel}(\boldsymbol s_{t-1})$ by incorporating the new information contained in $\boldsymbol a_t$ and $\boldsymbol o_t$. From assuming conditional independence between actions and observations given the state follows that Bayes filters update the belief in two steps: 1) \emph{prediction} using action $\boldsymbol a_{t}$ and 2) \emph{measurement update} using observation $\boldsymbol o_t$. 

1) The \emph{prediction} step is based on the \emph{motion model} $p(\boldsymbol s_t\mid \boldsymbol s_{t-1}, \boldsymbol a_{t})$, which defines how likely the robot enters state $\boldsymbol s_t$ if it performs action $\boldsymbol a_{t}$ in $\boldsymbol s_{t-1}$. Using the motion model, this step computes the \emph{predicted belief} $\overline{\text{bel}}(\boldsymbol s_t)$ by summing over all $\boldsymbol s_{t-1}$ from which $\boldsymbol a_t$ could have led to $\boldsymbol s_t$. 
\begin{align}
\overline{\text{bel}}(\boldsymbol s_t) &= \int p(\boldsymbol s_t\mid \boldsymbol s_{t-1}, \boldsymbol a_{t})\, \text{bel}(\boldsymbol s_{t-1}) \, d \boldsymbol s_{t-1}.
\label{eq:pred}
\end{align}

2) The \emph{measurement update} uses the \emph{measurement model} $p(\boldsymbol o_t \mid \boldsymbol s_t)$, which defines the likelihood of an observation $\boldsymbol o_t$ given a state $\boldsymbol s_t$. Using this model and observation $o_t$, this step updates the belief using Bayes' rule (with normalization $\eta$),
\begin{align}
\text{bel}(\boldsymbol s_t) = \eta \, p(\boldsymbol o_t \mid \boldsymbol s_t) \, \overline{\text{bel}}(\boldsymbol s_t).
\label{eq:update}
\end{align}

Any implementation of the Bayes filter algorithm for a continuous state space must represent a continuous belief--and thereby approximate it. Different approximations correspond to different Bayes filter implementations, for example histogram filters, which represent the belief by a histogram, Kalman filters, which represent it by a Gaussian, or particle filters, which represent the belief by a set of particles \citep{thrun_probabilistic_2005}.

\begin{figure*}[t]
\centering
\vspace{-0.5cm}
\begin{minipage}{.58\textwidth}  
\subfloat[Prediction and measurement update; boxes represent models, colored boxes are learned \label{subfig:overview}]{
\includegraphics[width=\textwidth,valign=t]{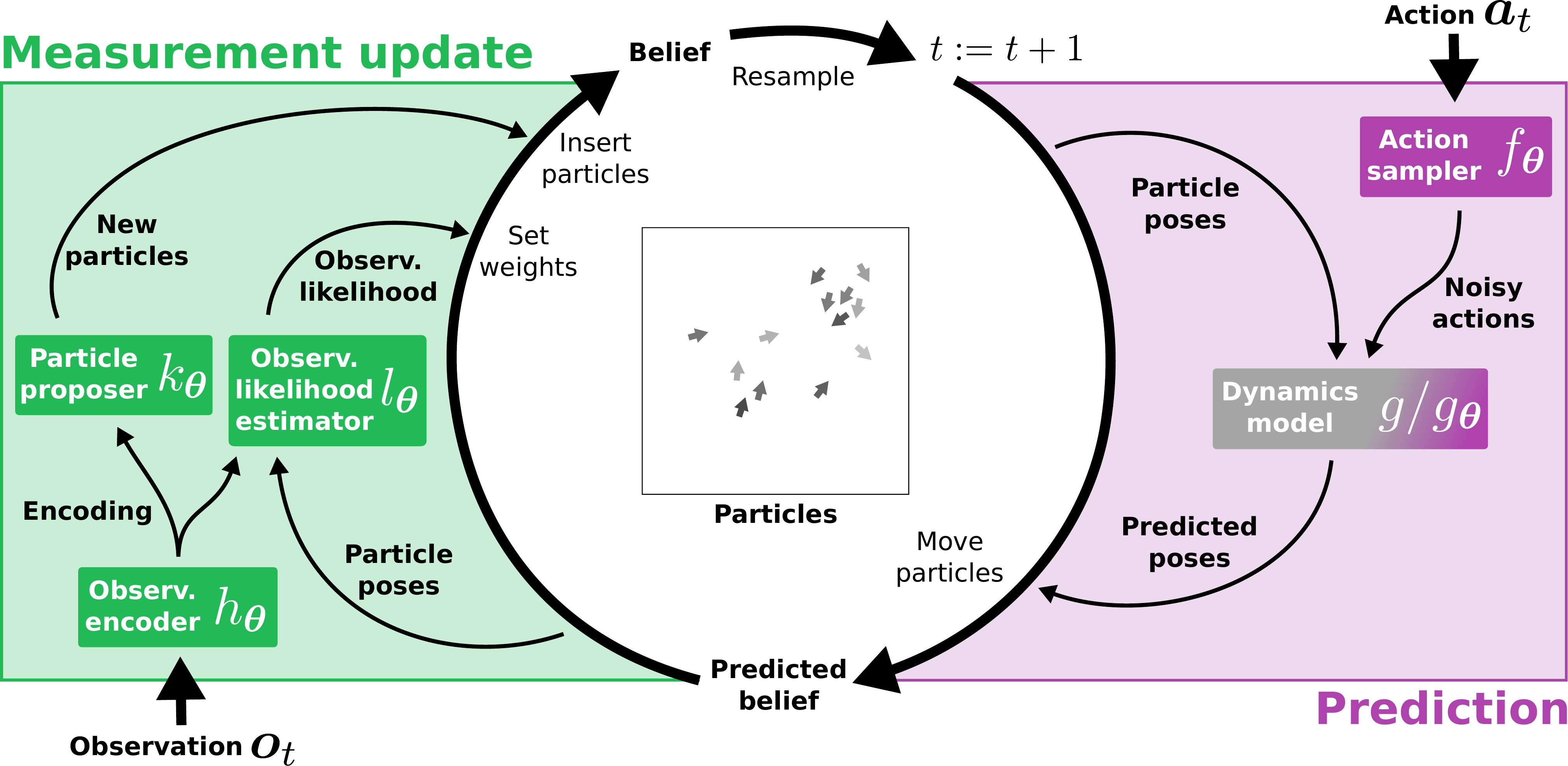}}
\end{minipage}
\hfill
\begin{minipage}{.40\textwidth}
%\subfloat[Proposal and resampling ratio over time\label{subfig:ratio}]{
%\includegraphics[width=\textwidth,valign=t]{images/ratio.pdf}}
%\\
\vspace{0.35cm}
\subfloat[Computing the gradient for end-to-end learning requires density estimation from the predicted particles (gray circles, darkness corresponds to particle weight). After converting the particles into a mixture of Gaussians (blue), we can compute the belief at the true state (orange bar at red x) and maximize it.\label{subfig:gaussians}]{
\includegraphics[width=\textwidth,valign=t]{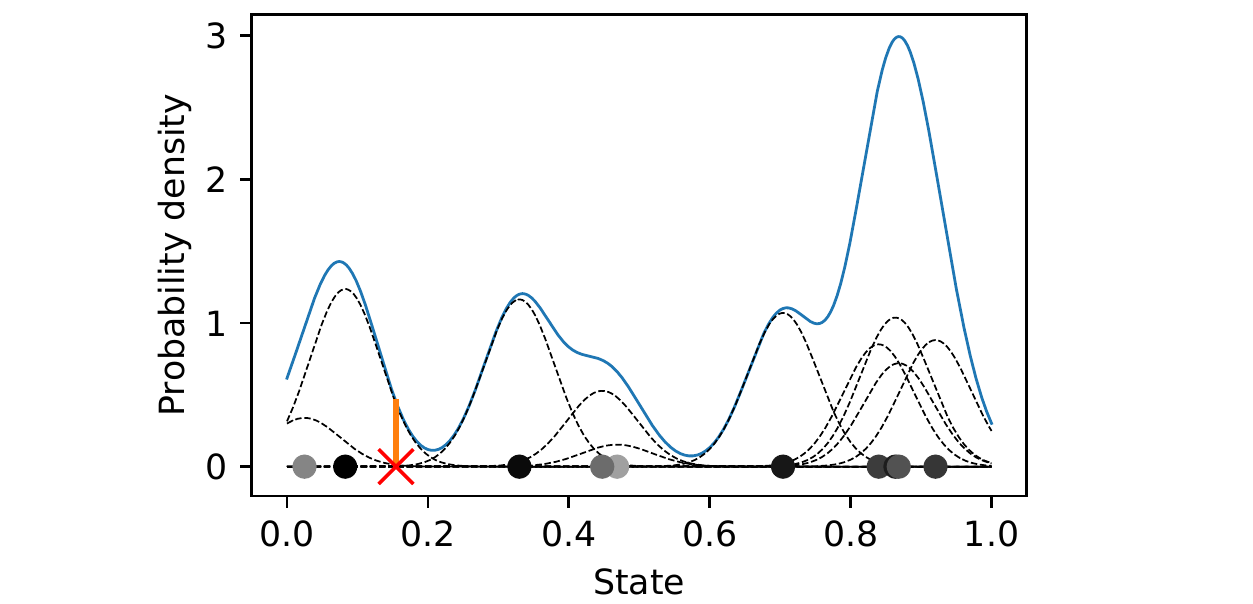}}
\end{minipage}
\caption{{\bf DPF overview}. Models in (a) can be learned end-to-end by maximizing the belief of the true state (b).}
\label{fig:pf}
\end{figure*}

% Differentiable particle filter with prediction, measurement update, and (non-differentiable) resampling. The measurement update uses a learned observation encoder, particle proposer, and observation likelihood estimator. The prediction step uses a learned action sampler. All learned models are feedforward networks. The dynamics model is given here, but can be learned as well, which we verify in our experiments.

%-------------------------------------------------------------------------------
\subsection{Particle Filters}
%-------------------------------------------------------------------------------

Particle filters approximate the belief with particles (or samples) $\mathcal{S}_t = \boldsymbol{s}^{[1]}_t, \boldsymbol{s}^{[2]}_t, \dots, \boldsymbol{s}^{[n]}_t$ with weights $w^{[1]}_t, w^{[2]}_t, \dots, w^{[n]}_t$. 
%Particle density represents probability---a concentration of particles corresponds to high probability in that region, a lack of particles corresponds to low probability. 
The particle filter updates this distribution by moving particles, changing their weights, and resampling them, which duplicates or removes particles proportionally to their weight. Resampling makes this Bayes filter implementation efficient by focusing the belief approximation on probable states.

The particle filter implements the prediction step (Eq.~\ref{eq:pred}) by moving each particle stochastically, which is achieved by sampling from a generative motion model,
\begin{align}
\forall_i: \;\; \boldsymbol{s}^{[i]}_{t} \sim p(\boldsymbol s_{t} \mid \boldsymbol a_{t}, \boldsymbol{s}^{[i]}_{t-1}).
\label{eq:pred_pf}
\end{align}
The particle filter implements the measurement update (Eq.~\ref{eq:update}) by setting the weight of each particle to the observation likelihood---the probability of the current observation conditioned on the state represented by the particle,
\begin{align}
\forall_i: \;\; w^{[i]}_{t} = p(\boldsymbol o_{t} \mid \boldsymbol{s}^{[i]}_{t}).
\label{eq:update_pf}
\end{align}
The particle set is then resampled by randomly drawing particles $\boldsymbol{s}^{[i]}_{t}$ proportionally to their weight $w^{[i]}_{t}$ before the filter performs the next iteration of prediction and update.

\section{Differentiable Particle Filters}
%======================================================================

Differentiable particle filters (DPFs) are a differentiable implementation of the particle filter algorithm with end-to-end learnable models. We  can also view DPFs as a new recurrent network architecture that encodes the algorithmic prior from particle filters in the network structure (see Fig.~\ref{subfig:overview}).

With end-to-end learning, we do not mean that every part of a system is learned but that the objective for the learnable parts is end-to-end performance. For efficient end-to-end learning in particle filters, we need learnable models and the ability to backpropagate the gradient through the particle filter algorithm---not to change the algorithm but to compute how to change the models to improve the algorithm's output. %To enable backpropagation, models and algorithm need to be differentiable. 

This section describes our DPF implementation. Our source code based on TensorFlow~\citep{tensorflow-15} and Sonnet~\citep{deepmind_sonnet:_2017} is available at \mbox{\url{https://github.com/tu-rbo/differentiable-particle-filters}}.

%-------------------------------------------------------------------------------
\subsection{Belief}
%-------------------------------------------------------------------------------

%We implement the set of $n$ weighted particles by a vector $\boldsymbol{w} \in \mathbb{R}^n$ and a matrix $S \in \mathbb{R}^{n\times d}$, where $d$ is the dimensionality of the state space. 

DPFs represent the belief at time $t$ by a set of weighted particles, $\text{bel}(\boldsymbol s_t)=(S_t, \boldsymbol{w}_t)$, where $S \in \mathbb{R}^{n\times d}$ describes $n$ particles in $d$-dimensional state space with weights $\boldsymbol{w} \in \mathbb{R}^n$. At every time step, DPFs update the previous belief $\text{bel}(\boldsymbol s_{t-1})$ with action $\boldsymbol{a}_{t}$ and observation $\boldsymbol{o}_{t}$ to get $\text{bel}(\boldsymbol s_t)$ (see Fig.~\ref{subfig:overview}).

%The particle positions and weights taken together represent the belief $\boldsymbol{b}_t=(S_t, \boldsymbol{w}_t)$ at time step $t$. 

%The filter generates the initial particle set by randomly sampling $n$ of the state labels from training data. As a result, the initial particle set approximates the state distribution observed during training. 

%-------------------------------------------------------------------------------
\subsection{Prediction}
%-------------------------------------------------------------------------------

The prediction step moves each particle by sampling from a probabilistic motion model (Eq.~\ref{eq:pred_pf}). Motion models often assume deterministic environments; they account for uncertainty by generating noisy versions of the commanded or measured action such that a different version of the action is applied to each particle \citep[chap.~5]{thrun_probabilistic_2005}. We follow the same approach by splitting the motion model into an \emph{action sampler}~$f$, which creates a noisy action $\hat{\boldsymbol{a}}^{[i]}$ per particle, and a \emph{dynamics model}~$g$, which moves each particle $i$ according to $\hat{\boldsymbol{a}}^{[i]}$.
\begin{align}
\hat{\boldsymbol{a}}^{[i]}_t &= \boldsymbol{a}_t + f_{\boldsymbol{\theta}}(\boldsymbol{a}_t, \boldsymbol{\epsilon}^{[i]} \sim \mathcal{N}), \label{eq:f}\\
\boldsymbol{s}^{[i]}_t &= \boldsymbol{s}^{[i]}_{t-1} + g(\boldsymbol{s}^{[i]}_{t-1}, \hat{\boldsymbol{a}}^{[i]}_t), \label{eq:g}
\end{align} 
%where $f_{\boldsymbol\theta}$ is a feedforward network, $\boldsymbol{\theta}$ are all parameters of the DPF, and $\boldsymbol\epsilon^{[i]}\in\mathbb{R}^{d}$ is a noise vector drawn from a standard normal distribution. $f_{\boldsymbol\theta}$ learns to sample from action-dependent motion noise and $g$ simulates how the actions change the state. If we know the underlying dynamics model we can directly implement its equations in $g$, which we do for our experiments. Alternatively, we could replace $g$ by a feedforward network $g_{\boldsymbol\theta}$ and learn the dynamics from data (see Table~\ref{tab:networks}).
where $f_{\boldsymbol\theta}$ is a feedforward network (see Table~\ref{tab:networks}), $\boldsymbol{\theta}$ are all parameters of the DPF, and $\boldsymbol\epsilon^{[i]}\in\mathbb{R}^{d}$ is a noise vector drawn from a standard normal distribution. Using the noise vector as input for a learnable generative model is known as the reparameterization trick~\citep{kingma2013auto}.
Here, this trick enables $f_{\boldsymbol\theta}$ to learn to sample from action-dependent motion noise. The resulting noisy actions are fed into $g$, which simulates how these actions change the state. Since we often know the underlying dynamics model, we can implement its equations in $g$. Alternatively, we can replace $g$ by a feedforward network $g_{\boldsymbol\theta}$ and learn the dynamics from data (tested in Section~\ref{sec:alg_priors_improve_perf}).

%-------------------------------------------------------------------------------
\subsection{Measurement Update}
\label{sec_measurement_update}
%-------------------------------------------------------------------------------

The measurement update uses the observation to compute particle weights (Eq.~\ref{eq:update_pf}). DPFs implement this update and additionally use the observation to propose new particles (see Fig.~\ref{subfig:overview}). The DPF measurement model consists of three components: a shared \emph{observation encoder}~$h$, which encodes an observation $\boldsymbol o_t$ into a vector $\boldsymbol e_t$, a \emph{particle proposer}~$k$, which generates new particles, and an \emph{observation likelihood estimator}~$l$, which weights each particle based on the observation.
\begin{align}
\boldsymbol e_t &= h_{\boldsymbol\theta}(\boldsymbol o_t), \label{eq:h}\\
\boldsymbol s^{[i]}_t &= k_{\boldsymbol\theta}(\boldsymbol{e}_t, \boldsymbol\delta^{[i]}\sim B), \label{eq:k}\\
w^{[i]}_t &= l_{\boldsymbol\theta}(\boldsymbol e_t, \boldsymbol{s}^{[i]}_t), \label{eq:l}
\end{align}
where $h_{\boldsymbol\theta}$, $k_{\boldsymbol\theta}$, and $l_{\boldsymbol\theta}$ are feedforward networks based on parameters $\boldsymbol\theta$; the input $\boldsymbol\delta^{[i]}$ is a dropout vector sampled from a Bernoulli distribution. Here, dropout is not used for regularization but as a source of randomness for sampling different particles from the same encoding $\boldsymbol e_t$ (see Table~\ref{tab:networks}).

\newcommand\Tstrut{\rule{0pt}{2.6ex}}         % = `top' strut
\newcommand\Bstrut{\rule[-0.9ex]{0pt}{0pt}}   % = `bottom' strut

%-------------------------------------------------------------------------------
\subsection{Particle Proposal and Resampling} 
%-------------------------------------------------------------------------------

We do \emph{not} initialize DPFs by uniformly sampling the state space---this would produce too few initial particles near the true state. Instead, we initialize DPFs by proposing particles from the current observation (as described above) for the first steps. During filtering, DPFs move gradually from particle proposal, which generates hypotheses, to resampling, which tracks and weeds out these hypotheses. The ratio of proposed to resampled particles follows an exponential function $\gamma^{t-1}$, where $\gamma$ is a hyperparameter set to $0.7$ in our experiments. We use 1000 particles for testing and 100 particles for training (to speed up the training process). DPFs implement resampling by stochastic universal sampling \citep{baker_reducing_1987}, which is not differentiable and leads to limitations discussed in Section \ref{sec:limitations}.

%DPFs combine two processes that generate particles: the first one proposes particles from the current observation (as described above) and the second one resamples existing particles based on their weight. During filtering, DPFs move gradually from the first process, which generates possible hypotheses, to the second, which tracks and weeds out these hypotheses. The ratio of proposed particles follows an exponential function $\gamma^{t-1}$ with hyperparameter $\gamma=0.7$ in our experiments. 

%-------------------------------------------------------------------------------
\subsection{Supervised Learning} 
\label{subsec:supl}
%-------------------------------------------------------------------------------

DPF models can be learned from sequences of supervised data $\boldsymbol{o}_{1:T}$, $\boldsymbol{a}_{1:T}$, $\boldsymbol{s}^*_{1:T}$ using maximum likelihood estimation by maximizing the belief at the \emph{true state} $\boldsymbol{s}^*_t$. To estimate $\text{bel}(\boldsymbol{s}^*_t)$ from a set of particles, we treat each particle as a Gaussian in a mixture model with weights $\boldsymbol{w}_t$ (see Fig.~\ref{subfig:gaussians}). For a sensible metric across state dimensions, we scale each dimension by dividing by the average step size $\text{E}_t[\text{abs}(\boldsymbol{s}^*_{t} - \boldsymbol{s}^*_{t-1})]$. This density estimation enables individual and end-to-end learning.

%-------------------------------------------------------------------------------
\subsubsection{Individual learning of the motion model} 
%-------------------------------------------------------------------------------

We optimize the motion model individually to match the observed motion noise by sampling states $\boldsymbol{s}^{[i]}_t$ from $\boldsymbol{s}^*_{t-1}$ and $\boldsymbol{a}_t$ using Eq.~\ref{eq:f}-\ref{eq:g} and maximizing the data likelihood as described above, $\boldsymbol{\theta}^*_f = \text{argmin}_{\boldsymbol{\theta}_f}-\log p(\boldsymbol{s}^*_t \mid \boldsymbol{s}^*_{t-1}, \boldsymbol{a}_{t}; \boldsymbol{\theta}_f)$. If the dynamics model $g$ is unknown, we train $g_{\boldsymbol\theta}$ by minimizing mean squared error between $g(\boldsymbol{s}^*_{t-1}, \boldsymbol{a}_t)$ and $\boldsymbol{s}^{*}_t - \boldsymbol{s}^{*}_{t-1}$.

%-------------------------------------------------------------------------------
\subsubsection{Individual learning of the measurement model} 
%-------------------------------------------------------------------------------

The particle proposer $k_{\boldsymbol{\theta}}$ is trained by sampling $\boldsymbol{s}^{[i]}_t$ from $\boldsymbol{o}_t$ using Eq.~\ref{eq:h}-\ref{eq:k} and maximizing the Gaussian mixture at $\boldsymbol{s}^*_t$. 

We train the observation likelihood estimator $l_{\boldsymbol \theta}$ (and $h_{\boldsymbol \theta}$) by maximizing the likelihood of observations in their state and minimizing their likelihood in other states, $\boldsymbol{\theta}^*_{h,l} = \text{argmin}_{\boldsymbol{\theta}_{h,l}}$ $ - \log(\text{E}_t[l_{\boldsymbol\theta}(h_{\boldsymbol\theta}(\boldsymbol o_t), \boldsymbol{s}^{*}_t)])- \log(1 - \text{E}_{t_1, t_2}[l_{\boldsymbol\theta}(h_{\boldsymbol\theta}(\boldsymbol o_{t_1}), \boldsymbol{s}^{*}_{t_2})]).$

%-------------------------------------------------------------------------------
\subsubsection{End-to-end learning}
%-------------------------------------------------------------------------------

For end-to-end learning, we apply DPFs on overlapping subsequences and maximize the belief at all true states along the sequence as described above,
%\begin{align*}
$$\boldsymbol{\theta}^* = \text{argmin}_{\boldsymbol{\theta}} -\log \text{E}_t[\text{bel}(\boldsymbol{s}^*_t; \boldsymbol{\theta})].$$
%\end{align*}

%-------------------------------------------------------------------------------
\subsection{Limitations and Future Work}
\label{sec:limitations}
%-------------------------------------------------------------------------------

We compute the end-to-end gradient by backpropagation from the DPF output through the filtering loop. \emph{Since resampling is not differentiable, it stops the gradient computation after a single loop iteration.} Therefore, the gradient neglects the effects of previous prediction and update steps on the current belief. This limits the scope of our implementation to supervised learning, where predicting the Markov state at each time step is a useful objective that facilitates future predictions. Differentiable resampling could still improve supervised learning, e.g. by encouraging beliefs to overestimate uncertainty, which reduces performance at the current step but can potentially increase robustness of future state estimates. 

Since it is difficult to generate training data that include the true state $\boldsymbol{s}^*_t$ outside of simulation, we must work towards unsupervised learning, which will require backpropagation through multiple time steps because observations are generally non-Markov. Here are two possible implementations of differentiable resampling that could be the starting point of future work: a) Partial resampling: sample only $m$ particles in each step; keep $n-m$ particles from the previous time step; the gradient can flow backwards through those. b) Proxy gradients: define a proxy gradient for the weight of a resampled particle that is tied to the particle it was sampled from; the particle pose is already connected to the pose of the particle it was sampled from; the gradient can flow through these connections.

%Differentiable resampling and backpropagation through time in DPFs are important topics for future work.

%Here are two possible approaches to this problem: 

%===============================================================================
\section{Experiments}
\label{sec:experiments}
%===============================================================================

\begin{table}[t]
\caption{{\bf Feedforward networks for learnable DPF models}}\label{tab:networks}
\begin{tabular}{p{0.2cm} p{7.8cm}}
%\noalign{\smallskip}
\hline
$f_{\boldsymbol\theta}$: & 2 x fc(32, relu), fc(3) + mean centering across particles \Tstrut\\[2pt]
$g_{\boldsymbol\theta}$: & 3 x fc(128, relu), fc(3) + scaled by $\text{E}_t[\text{abs}(\boldsymbol{s}_{t} - \boldsymbol{s}_{t-1})]$\\[2pt]
$h_{\boldsymbol\theta}$: & conv(3x3, 16, stride 2, relu), conv(3x3, 32, stride 2, relu), conv(3x3, 64, stride 2, relu), dropout(keep 0.3), fc(128, relu)\\[2pt]
$k_{\boldsymbol\theta}$: & fc(128, relu), dropout*(keep 0.15), 3 x fc(128, relu), fc(4, tanh) \\[2pt]
$l_{\boldsymbol\theta}$: & 2 x fc(128, relu), fc(1, sigmoid scaled to range [0.004, 1.0]) \Bstrut\\
\hline
\end{tabular}
%\vspace{2cm}
\Tstrut
fc: fully connected, conv: convolution, *: applied at training and test time
\end{table}

We evaluated DPFs in two state estimation problems in robotics: \emph{global localization} and \emph{visual odometry}. We tested global localization in simulated 3D mazes based on vision and odometry. We focused on this task because it requires simultaneously considering multiple hypotheses, which is the main advantage of particle filters over Kalman filters. Here, we evaluated: a) the effect of end-to-end learning compared to individual learning and b) the influence of algorithmic priors encoded in DPFs by comparing to generic LSTMs. To show the versatility of DPFs and to compare to published results with backprop Kalman filters (BKFs) \citep{haarnoja_backprop_2016}, we also apply DPFs to the KITTI visual odometry task~\citep{geiger2013vision}. The goal is to track the pose of a driving car based on a first-person-view video. In both tasks, DPFs use the known dynamics model $g$ but do not assume any knowledge about the map of the environment and learn the measurement model entirely from data.

Our global localization results show that 1) algorithmic priors enable explainability, 2) end-to-end learning improves performance but sequencing individual and end-to-end learning is even more powerful, 3) algorithmic priors in DPFs improve performance compared to LSTMs reducing the error by $\sim$80\%, and 4) algorithmic priors lead to policy invariance: While the LSTM baseline learns localization in a way that stops working when the robot behaves differently ($\sim$84\% error rate), localization with the DPF remains useful with different policies ($\sim$15\% error rate).

In the visual odometry task, DPFs outperform BKFs even though the task exactly fits the capabilities and limitations of Kalman filters---tracking a unimodal belief from a known initial state. This result demonstrates the applicability of DPFs to tasks with different properties: higher frequency, longer sequences, a 5D state instead of a 3D state, and latent actions. The result also shows that DPFs work on real data and are able to learn measurement models that work for visually diverse observations based on less than 40 minutes of video. 

%-------------------------------------------------------------------------------
\subsection{Global Localization Task}
\label{subsec:problem_setting}
%-------------------------------------------------------------------------------

\begin{figure}[t]
    \centering
    \subfloat[Maze 1 (10x5)]{
		\includegraphics[width=0.215\columnwidth]{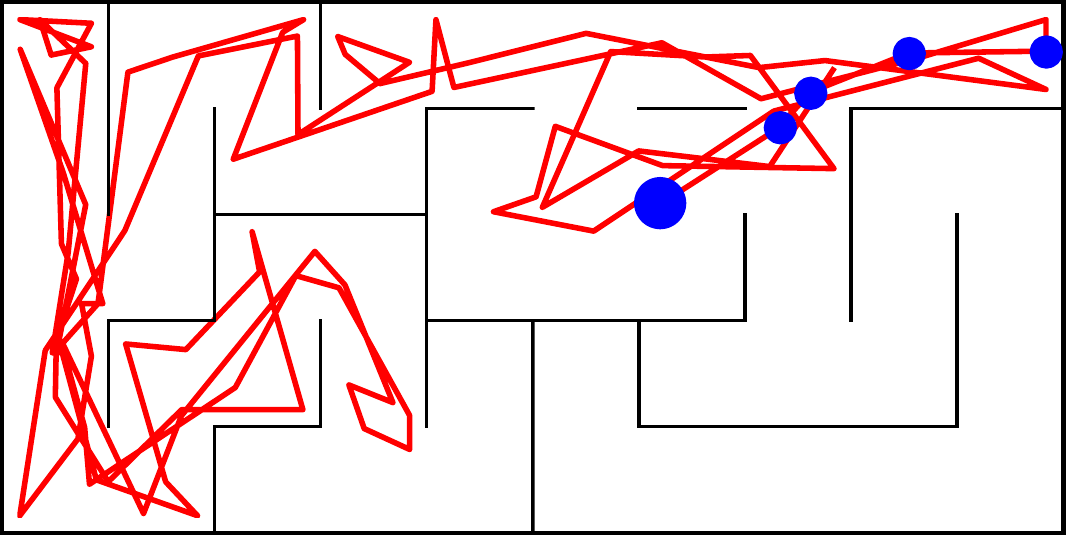}\label{fig:maze1}} % 
	% 0.21
    \hfill
    \subfloat[Maze 2 (15x9)]{
		\includegraphics[width=0.3175\columnwidth]{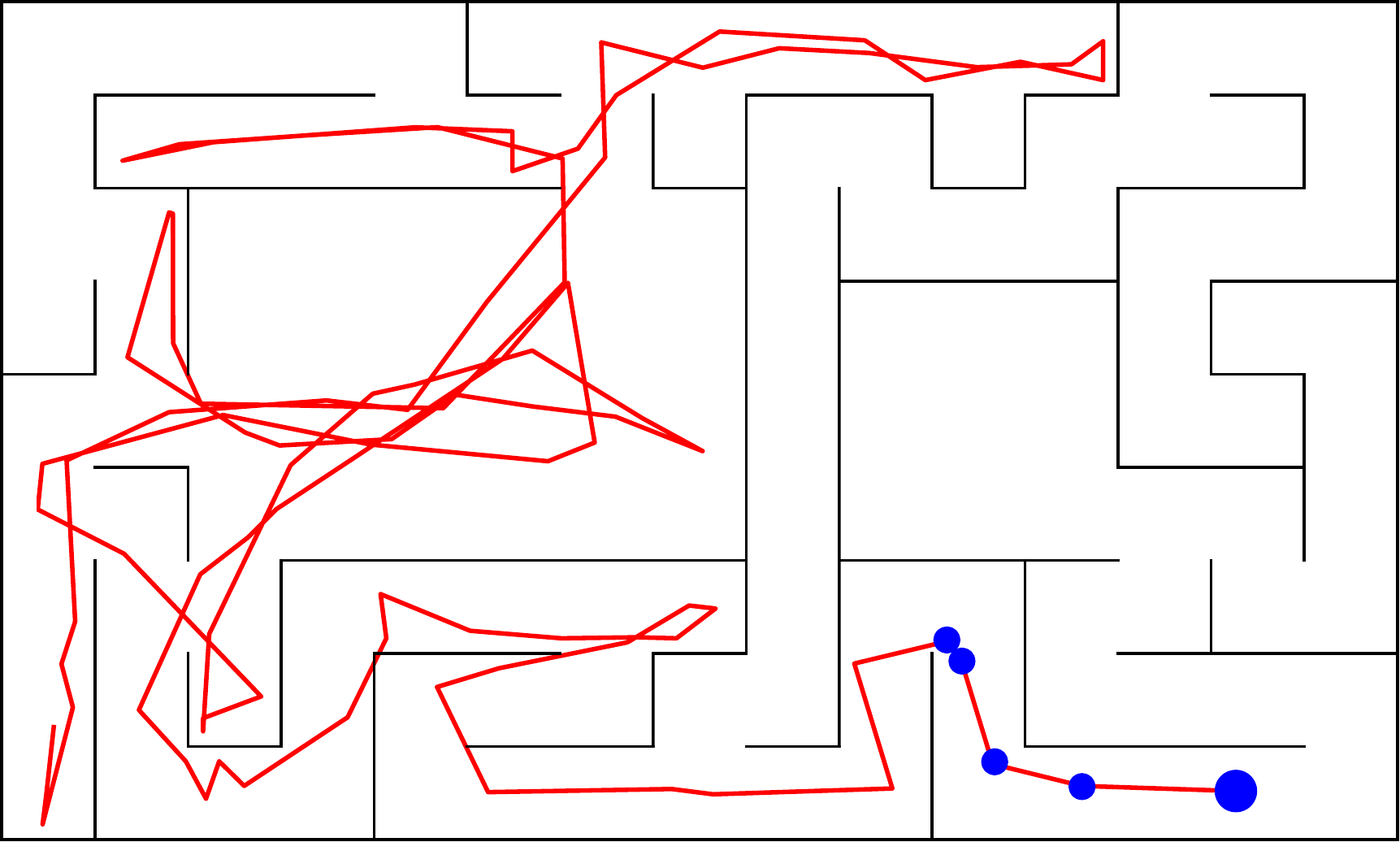}\label{fig:maze2}} %
	% 0.5292
   \hfill
   \subfloat[Maze 3 (20x13)]{
   		\includegraphics[width=0.425\columnwidth]{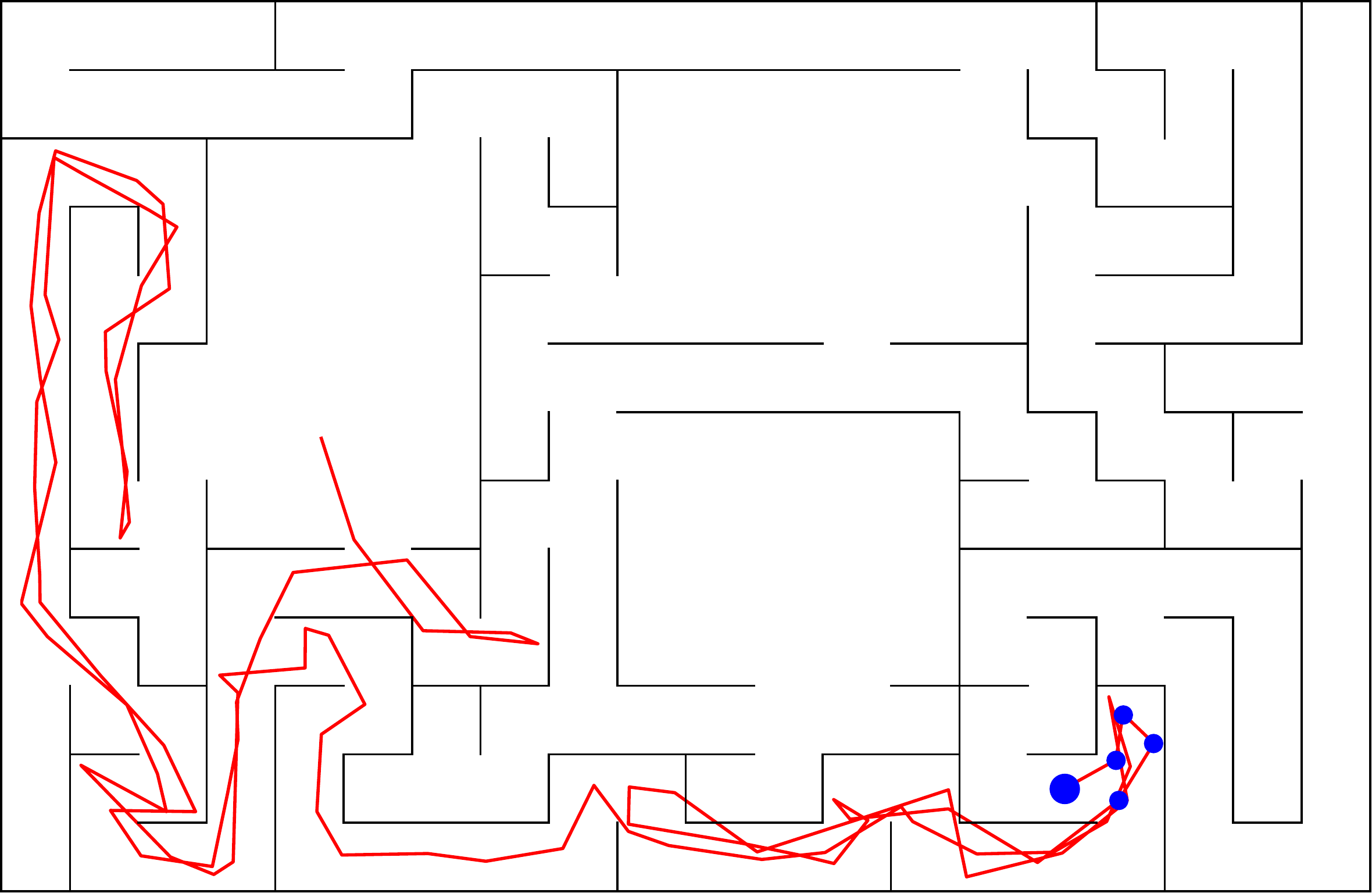}\label{fig:maze3}}
    \\
   \subfloat[Maze 1 observations]{
		\includegraphics[width=0.3\columnwidth]{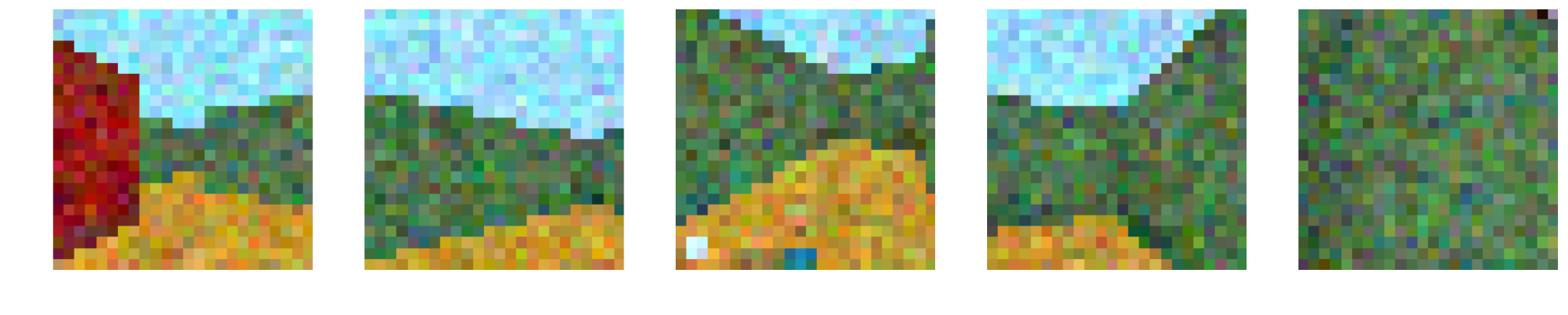}\label{fig:maze1_obs}}
    \hfill
    \subfloat[Maze 2 observations]{
		\includegraphics[width=0.3\columnwidth]{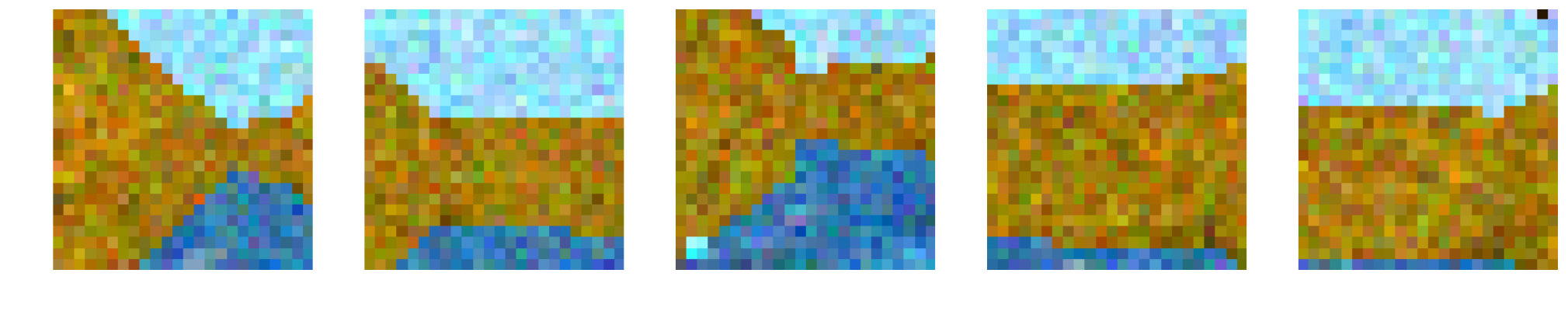}\label{fig:maze2_obs}}
   \hfill
  \subfloat[Maze 3 observations]{
	\includegraphics[width=0.3\columnwidth]{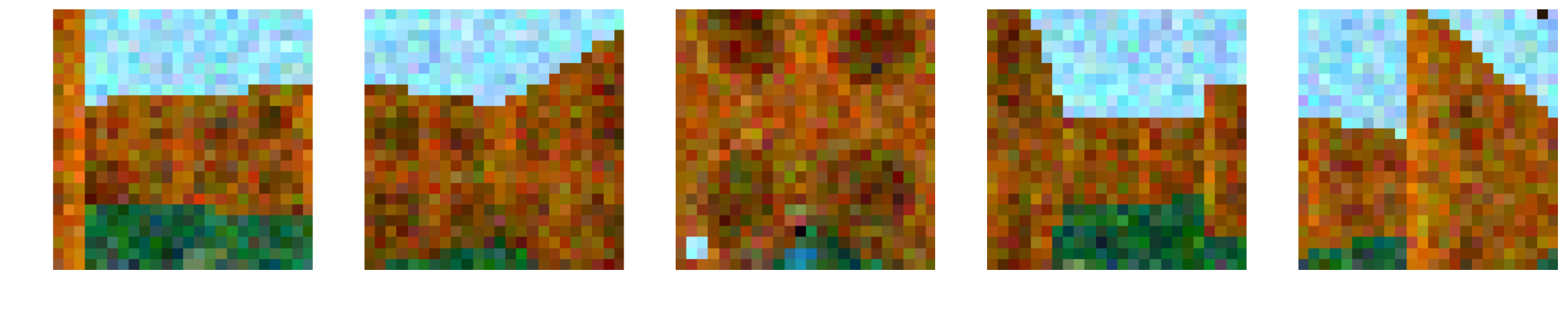}\label{fig:maze3_obs}}
    \caption{{\bf Three maze environments}. Red lines show example trajectories of length 100. Blue circles show the first five steps, of which the observations are depicted below.}
    \label{fig:tasks}
 
\end{figure}

The global localization task is about estimating the pose of a robot based on visual and odometry input. All experiments are performed in modified versions of the navigation environments from DeepMind Lab \citep{beattie_deepmind_2016}, where all objects and unique wall textures were removed to ensure partial observability. Data was collected by letting the simulated robot wander through the mazes (see Fig.~\ref{fig:tasks}). The robot followed a hand-coded policy that moves in directions with high depth values from RGB-D input and performs 10\% random actions. For each maze, we collected 1000 trajectories of 100 steps with one step per second for training and testing. As input for localization, we only used RGB images and odometry, both with random disturbances to make the task more realistic. For the observations, we randomly cropped the rendered $32\times32$ RGB images to $24\times24$ and added Gaussian noise ($\sigma=20$, see Fig.~\mbox{\ref{fig:tasks}d-f}). As actions, we used odometry information that corresponds to the change in position and orientation from the previous time step in the robot's local frame, corrupted with multiplicative Gaussian noise ($\sigma=0.1$). All methods were optimized on short trajectories of length 20 with Adam \citep{Kingma-14} and regularized using dropout \citep{srivastava2014dropout} and early stopping. We will now look at the results for this task.

%----------------------------
\subsubsection{\bf Algorithmic priors enable explainability}
%----------------------------

Due to the algorithmic priors in DPFs, the models remain explainable even after end-to-end learning. We can therefore examine a) the motion model, b) the measurement model, and c) their interplay during filtering. Unless indicated otherwise, all models were first learned individually and then end-to-end.

\paragraph{Motion Model}

\begin{figure}[t]
    \centering
    \subfloat[Predictions with learned motion model]{\includegraphics[width=0.57\columnwidth]{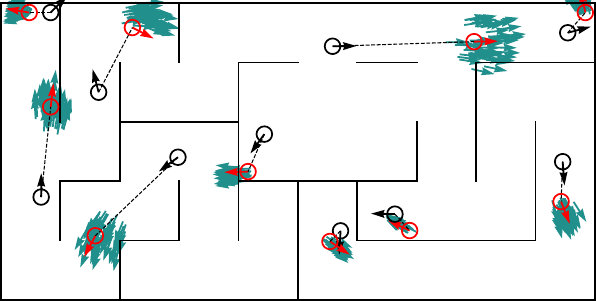}\label{subfig:motion_samples}} % 
    \hfill
    \subfloat[Comparison of learned noise]{\includegraphics[width=0.42\columnwidth]{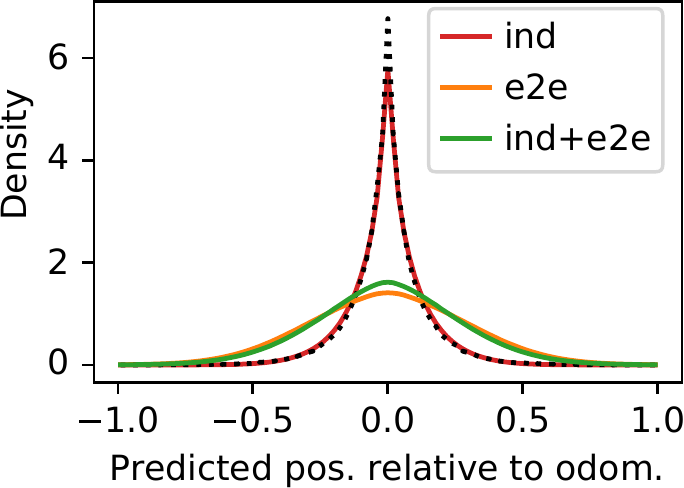}\label{subfig:motion_stats}}
    \caption{{\bf Learned motion model}. (a) shows predictions (cyan) of the state (red) from the previous state (black). (b) compares prediction uncertainty in x to true odometry noise (dotted line).}
    \label{fig:motion_model}
\end{figure}

Fig.~\ref{subfig:motion_samples} shows subsequent robot poses together with predictions from the motion model. These examples show that the model has learned to spread the particles proportionally to the amount of movement, assigning higher uncertainty to larger steps. But how does this behavior depend on whether the model was learned individually or end-to-end?

Fig.~\ref{subfig:motion_stats} compares the average prediction uncertainty using models from different learning schemes. The results show that individual learning produces an accurate model of the odometry noise (compare red and the dotted black lines). End-to-end learning generates models that overestimate the noise (green and orange lines), which matches insights of experts in state estimation who report that ``many of the models that have proven most successful in practical applications vastly overestimate the amount of uncertainty'' \cite[p.~118]{thrun_probabilistic_2005}.

\begin{figure}[t]
    \centering
    \captionsetup[subfloat]{farskip=2pt,captionskip=0pt}
    \subfloat[Obs.]{
		\includegraphics[width=0.05\textwidth]{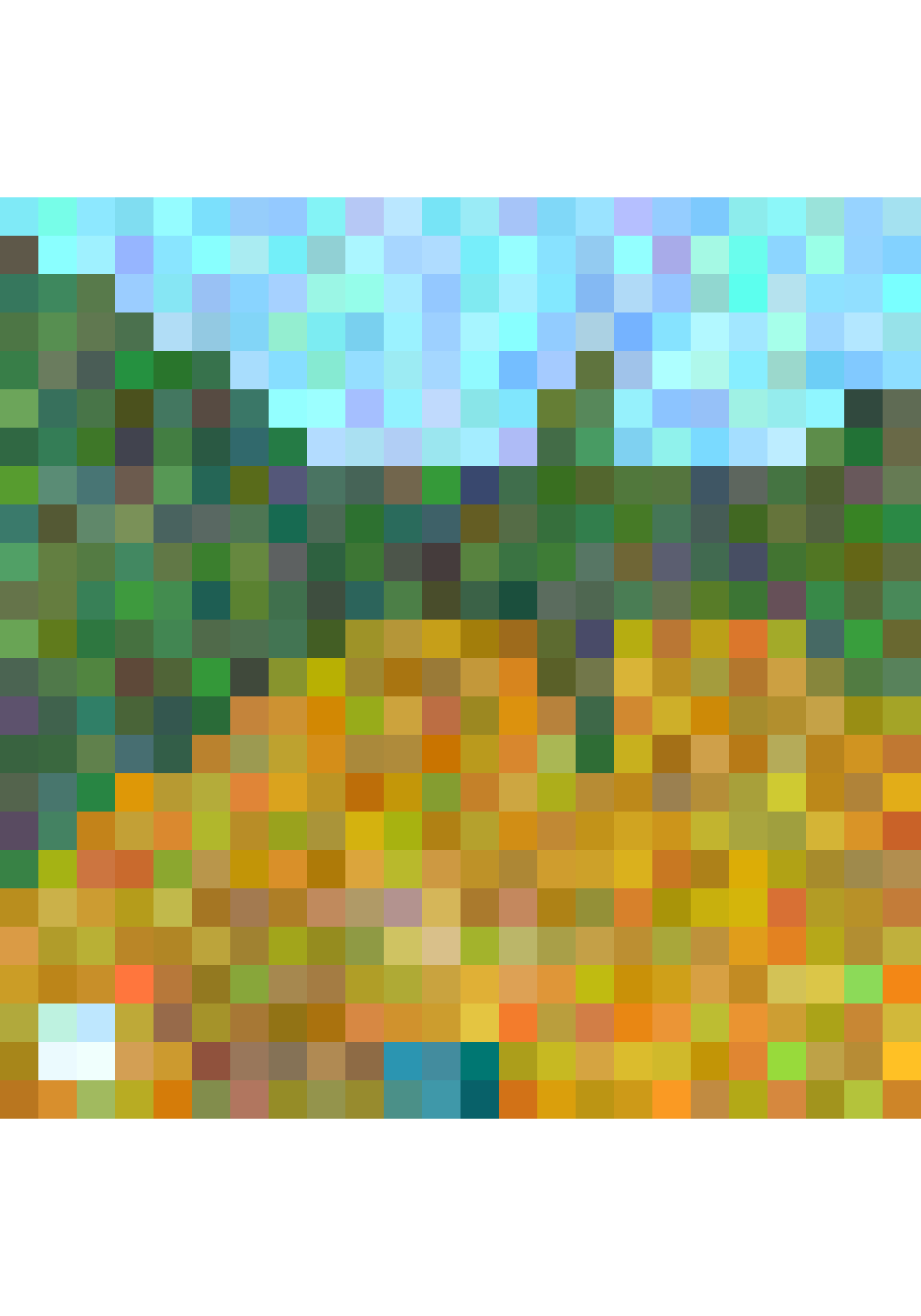}} % 
	\hfill
	\subfloat[Particle proposer]{
		\includegraphics[width=0.18\textwidth]{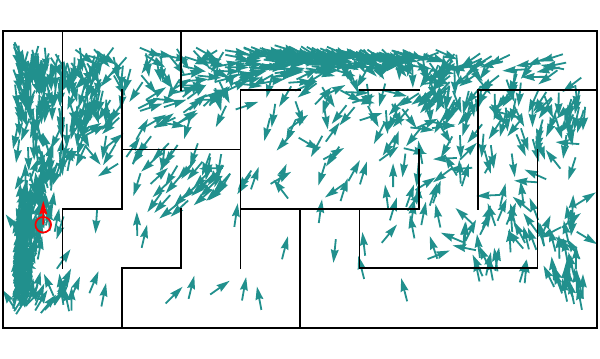}} % 
	\hfill
	\subfloat[Obs. likelihood estimator $\quad\quad$]{
		\includegraphics[width=0.233\textwidth]{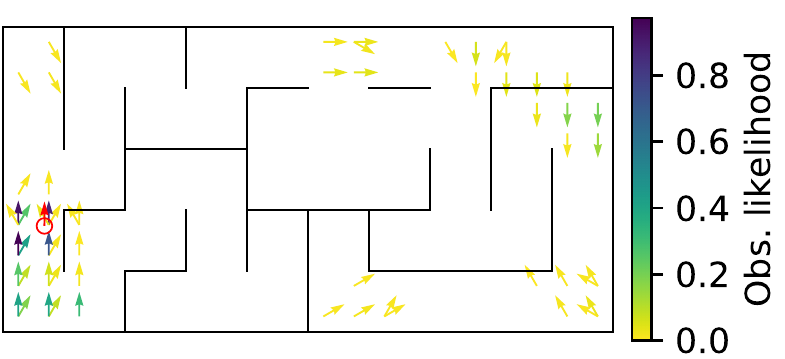}} % 
	\\
	\subfloat[Obs.]{
		\includegraphics[width=0.05\textwidth]{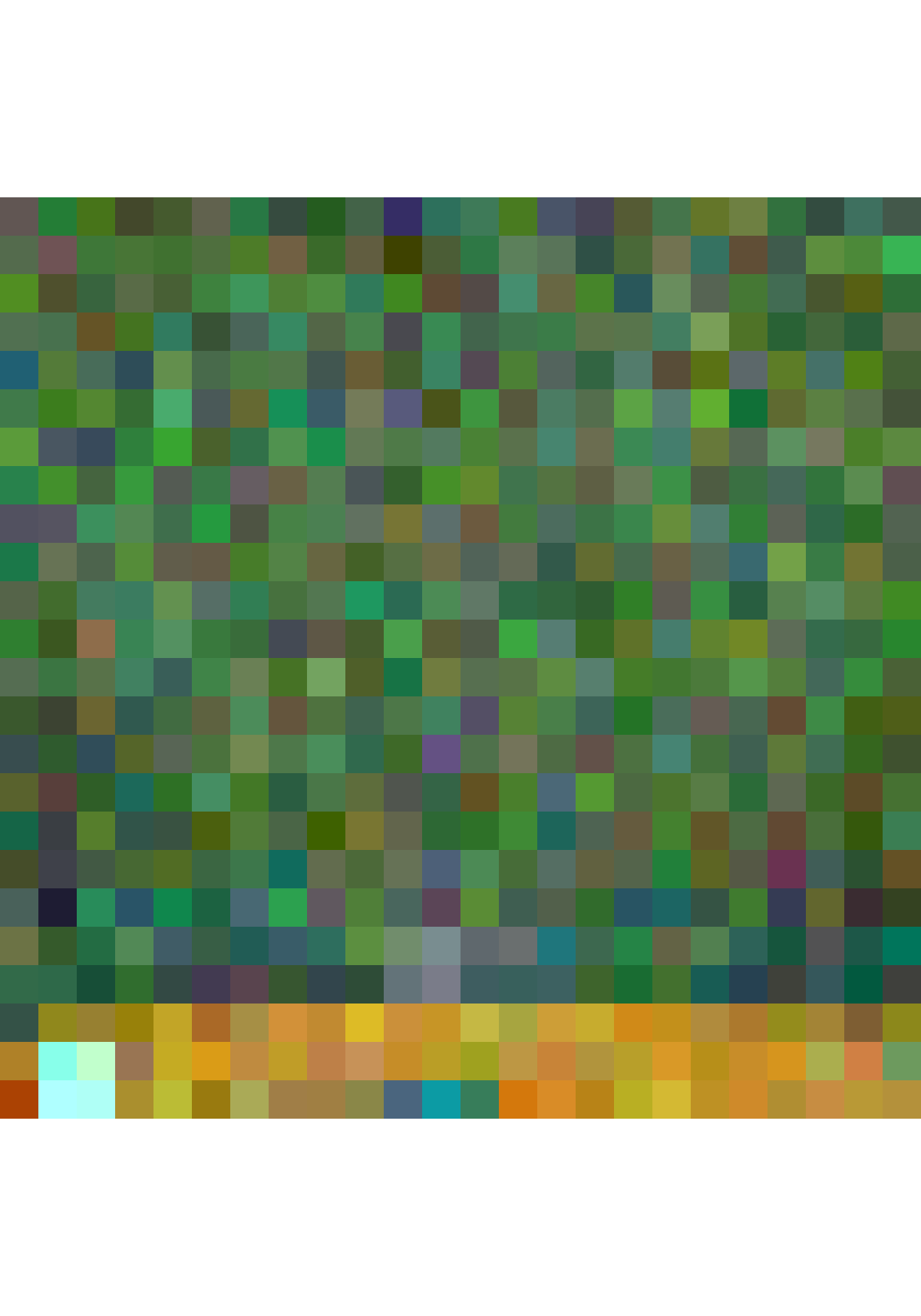}\label{subfig:obs2}} % 
	\hfill
	\subfloat[Particle proposer]{
		\includegraphics[width=0.18\textwidth]{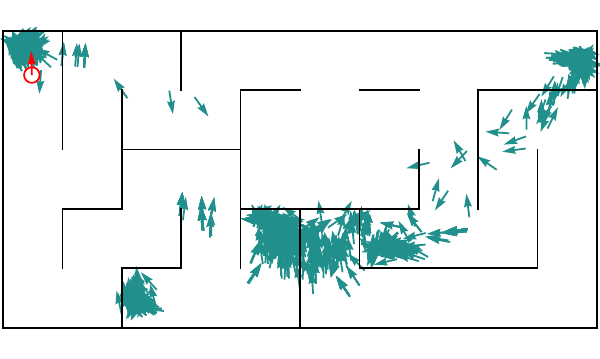}} % 
	\hfill
	\subfloat[Obs. likelihood estimator $\quad\quad$]{
		\includegraphics[width=0.233\textwidth]{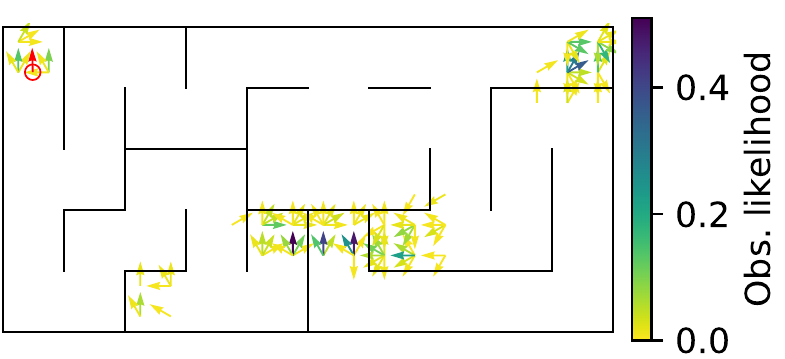}} % 
	\\
	\subfloat[Obs.]{
		\includegraphics[width=0.05\textwidth]{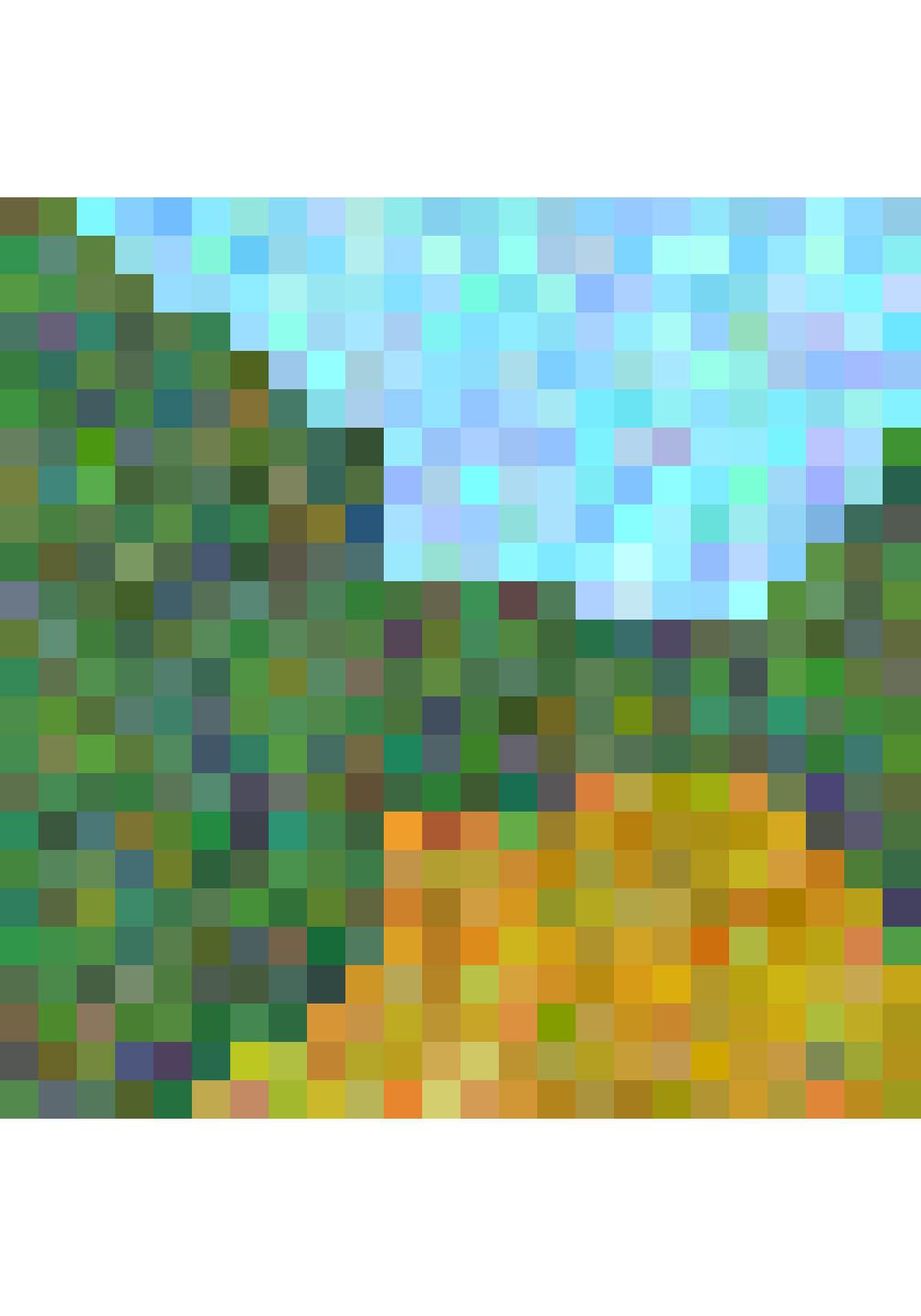}\label{subfig:obs3}} % 
	\hfill
	\subfloat[Particle proposer]{
		\includegraphics[width=0.18\textwidth]{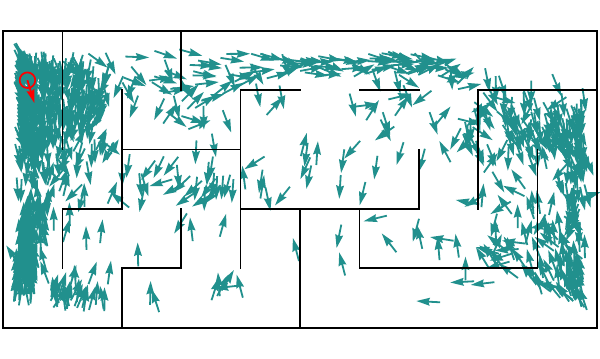}} % 
	\hfill
	\subfloat[Obs. likelihood estimator $\quad\quad$]{
		\includegraphics[width=0.233\textwidth]{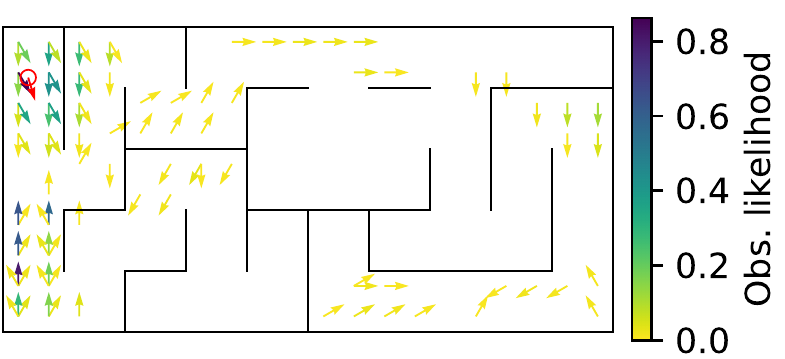}} % 
    \caption{{\bf Learned measurement model}. Observations, corresponding model output, and true state (red). To remove clutter, the observation likelihood only shows above average states.}
    \label{fig:measurement_model}
\end{figure}

\begin{figure*}[t]
\vspace{-0.1cm}
\centering
\includegraphics[width=\textwidth]{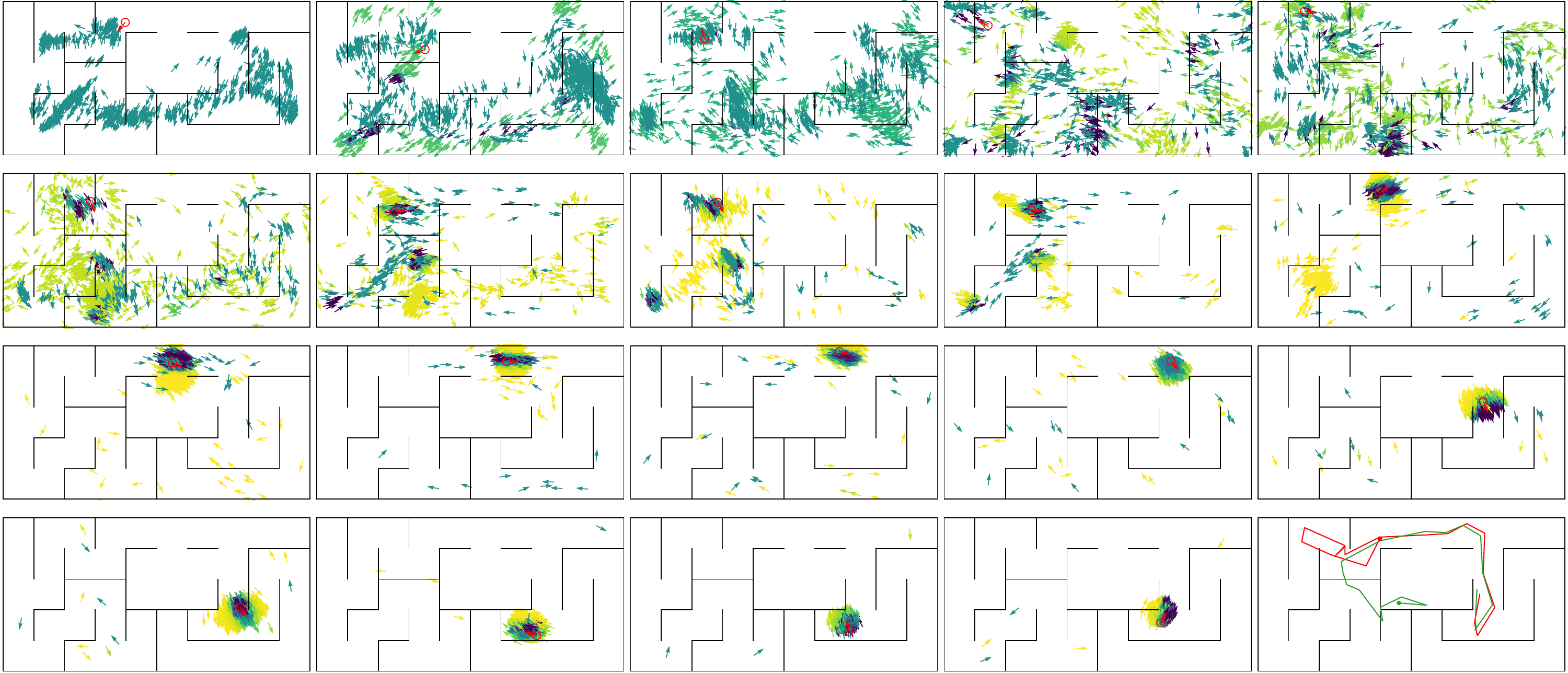} 
\\
\vspace{0.2cm}
\includegraphics[width=\textwidth]{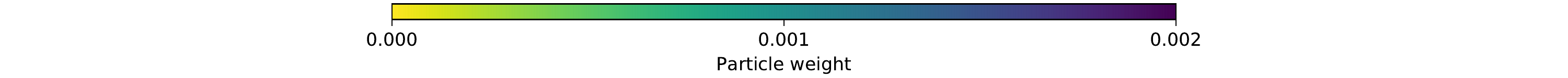}
\caption{{\bf Global localization with DPFs}. One plot per time step of a test trajectory: true state (red), 1000 particles (proposed particles have weight 0.001). Last plot: the weighted particle mean (green) matches the true state after the first few steps.}
\label{fig:qresult}
\end{figure*}

\paragraph{Measurement Model}

Fig.~\ref{fig:measurement_model} shows three example observations and the corresponding outputs of the measurement model: proposed particles and weights depending on particle position. Note how the model predicts particles and estimates high weights at the true state and other states in locally symmetric parts of the maze. We can also see that the data distribution shapes the learned models, e.g. by focusing on dead ends for the second observation, which is where the robot following the hand-coded policy will look straight at a wall before turning around. Similar to motion models, end-to-end learned measurement models are not accurate but effective for end-to-end state estimation, as we will see next.

\paragraph{Filtering}

Figure~\ref{fig:qresult} shows filtering with learned models. The DPF starts by generating many hypotheses (top row). Then, hypotheses form clusters and incorrect clusters vanish when they are inconsistent with observations (second row). Finally, the remaining cluster tracks the true state.

\begin{figure*}[t]
	\vspace{-0.1cm}
    \centering
    \subfloat[Maze 1 (10x5)]{
		\includegraphics[width=0.32\textwidth]{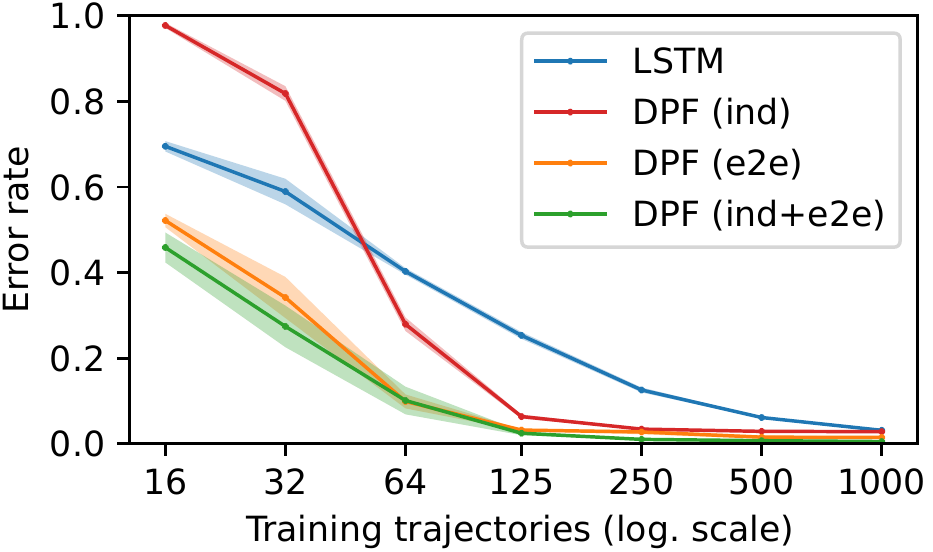}\label{subfig:lc1}}
    \hfill
    \subfloat[Maze 2 (15x9)]{
		\includegraphics[width=0.32\textwidth]{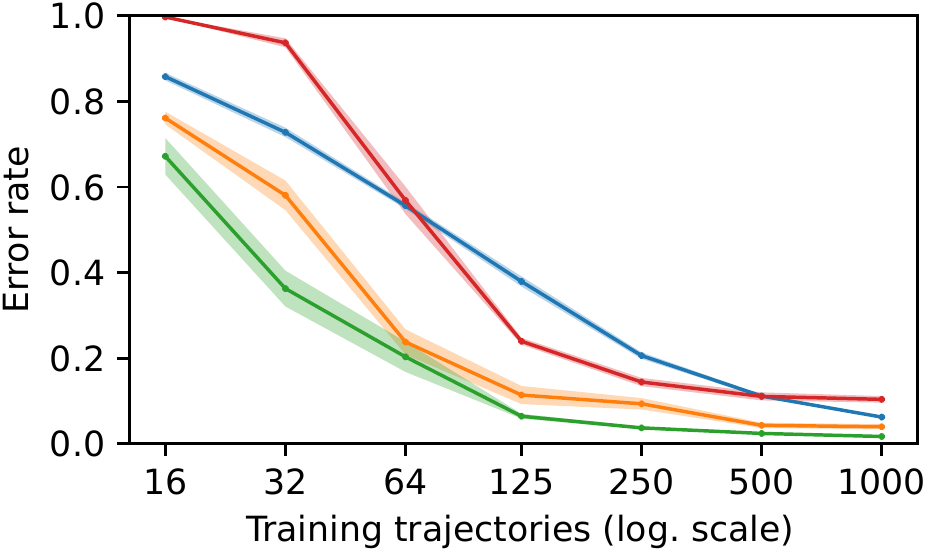}\label{subfig:lc2}}
    \hfill
    \subfloat[Maze 3 (20x13)]{
		\includegraphics[width=0.32\textwidth]{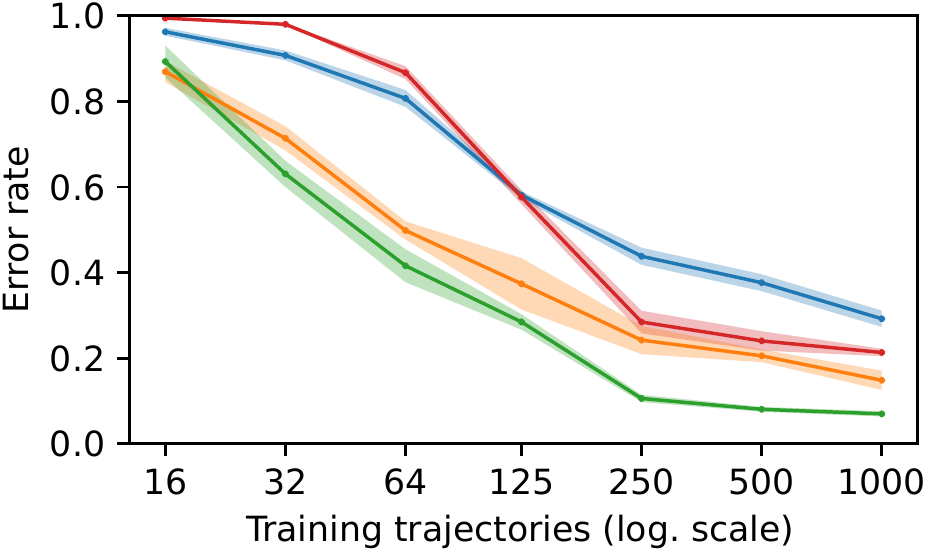}\label{subfig:lc3}}	
	\\
	\subfloat[Maze 1 (10x5), relative to LSTM]{
		\includegraphics[width=0.32\textwidth]{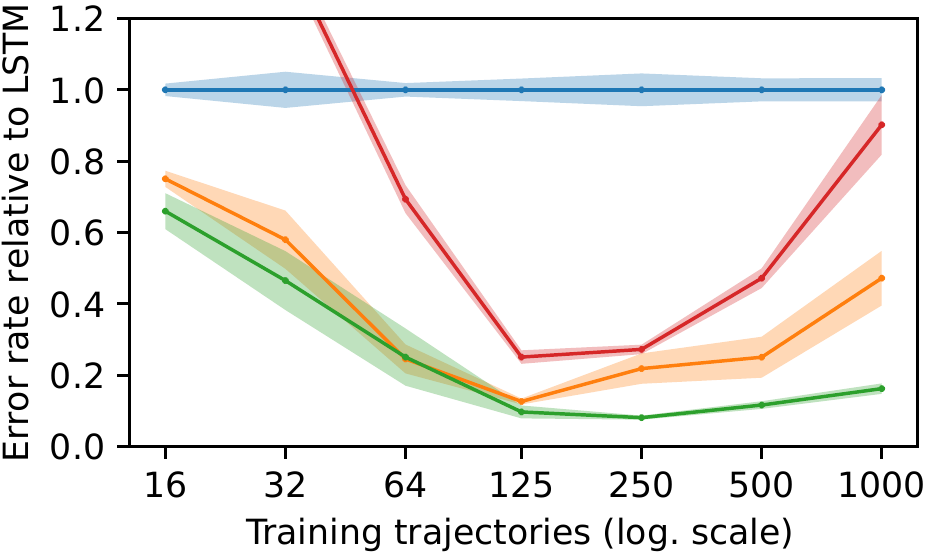}\label{subfig:nlc1}}
    \hfill
    \subfloat[Maze 2 (15x9), relative to LSTM]{
		\includegraphics[width=0.32\textwidth]{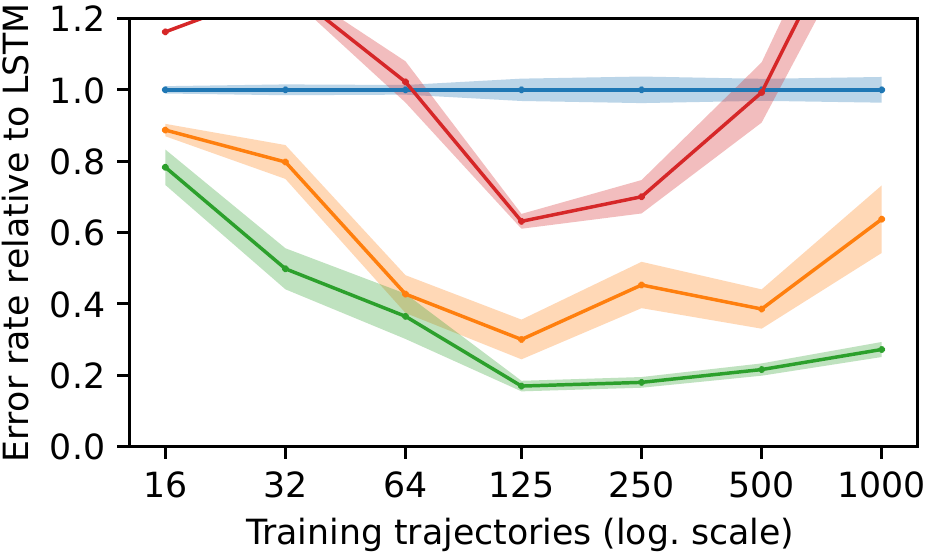}\label{subfig:nlc2}}
    \hfill
    \subfloat[Maze 3 (20x13), relative to LSTM]{
		\includegraphics[width=0.32\textwidth]{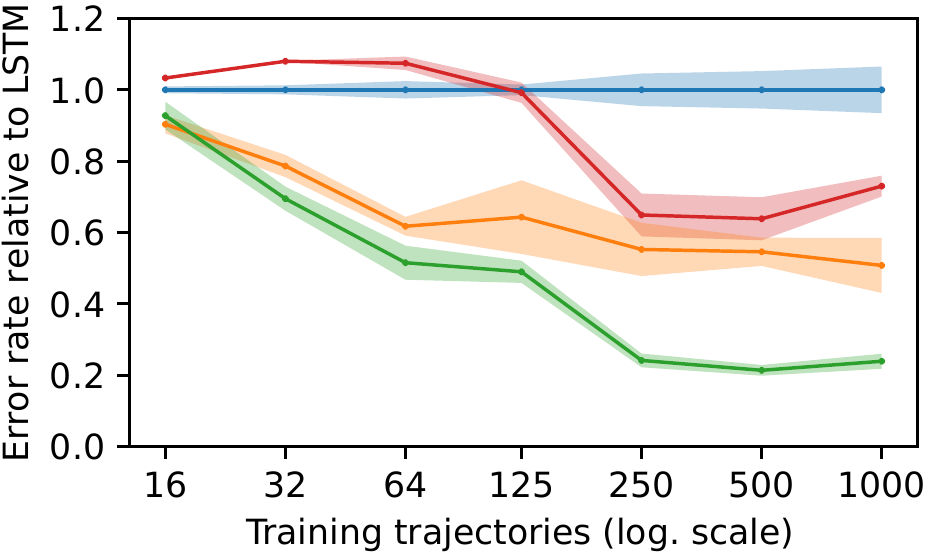}\label{subfig:nlc3}}	
	\caption{{\bf Learning curves} in all mazes (a-c), also relative to LSTM baseline (d-f). ind: individual learning, e2e: end-to-end learning. Shaded areas denote standard errors.
	}
	\label{fig:lc}
\end{figure*}

\begin{figure}[t]
\centering
\includegraphics[width=\columnwidth]{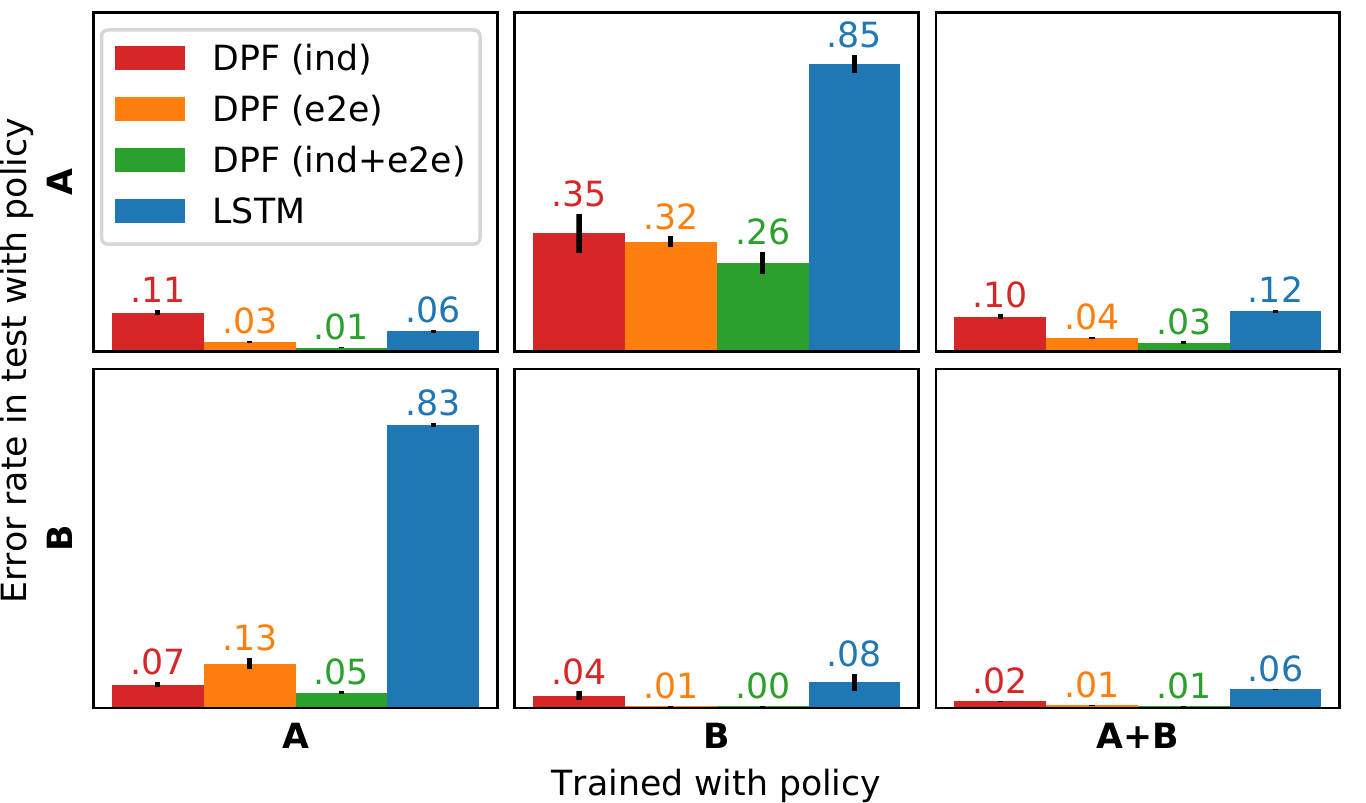}
\caption{{\bf Generalization between policies} in maze 2. A:~heuristic exploration policy, B:~shortest path policy. Methods were trained using 1000 trajectories from A, B, or an equal mix of A and B, and then tested with policy A or B. }
\label{fig:policies}
\end{figure}

%----------------------------
\subsubsection{\bf End-to-end learning improves performance}
%----------------------------

To quantify the effect of end-to-end learning on state estimation performance, we compared three different learning schemes for DPFs: individual learning of each model ({\bf ind}), end-to-end learning ({\bf e2e}), and both in sequence ({\bf ind+e2e}). We evaluated performance in all three mazes and varied the amount of training trajectories along a logarithmic scale from 32 to 1000. We measured localization performance by \emph{error rate}, where we consider a prediction erroneous if the distance to the true state, divided by  $\text{E}_t[\text{abs}(\boldsymbol{s}_{t} - \boldsymbol{s}_{t-1})]$, is greater than 1.
%the Euclidean distance to the true state is greater than one after scaling each state dimension by dividing by the average state change. 

The resulting learning curves in Fig.~\ref{fig:lc}a-c show that end-to-end learned DPFs (orange line) consistently outperform individually trained DPFs (red line) across all mazes. Individual training is worst with few training trajectories (less than 64) but also plateaus with more data (more than 125 trajectories). In both cases, the problem is that the models are not optimized for state estimation performance. With few data, training does not take into account how unavoidable model errors affect filtering performance. With lots of data, the models might be individually accurate but suboptimal for end-to-end filtering performance. End-to-end learning consistently leads to improved performance for the same reasons. 

Performance improves even more when we sequence individual and end-to-end learning (green line in Fig.~\ref{fig:lc}a-c). Individual pretraining helps because it incorporates additional information about the function of each model into the learning process, while end-to-end learning incorporates information about how these models affect end-to-end performance. Naturally, it is beneficial to combine both sources of information.

%----------------------------
\subsubsection{\bf Algorithmic priors improve performance}
\label{sec:alg_priors_improve_perf}
%----------------------------

To measure the effect of the algorithmic priors encoded in DPFs, we compare them with a generic neural network baseline that replaces the filtering loop with a two-layer long-short-term memory network (LSTM) \citep{hochreiter_long_1997}. The baseline architecture uses the same convolutional network architecture as the DPF---it embeds images using a convolutional network $h_{\boldsymbol\theta}$, concatenates the embedding with the action vector and feeds the result into 2xlstm(512), 2xfc(256, relu), and fc(3)---and is trained end-to-end to minimize mean squared error.% at all time steps.

The comparison between DPF (ind+e2e) and the LSTM baseline (blue) in Fig.~\ref{fig:lc}a-c shows that the error rate of DPF (ind+e2e) is lower than for LSTM for all mazes and all amounts of training data. Also in all mazes, DPF (ind+e2e) achieve the final performance of LSTM already with 125 trajectories, $\frac{1}{8}$ of the full training set. 

We performed a small ablation study in maze 2 to quantify the effect the known dynamics model on this performance. When the dynamics model is learned, the final error rate for DPFs increases from 1.6\% to 2.7\% compared to 6.0\% error rate for LSTMs. This shows that knowing the dynamics model is helpful but not essential for DPF's performance.

To visualize the performance relative to the baseline, we divided all learning curves by LSTM's performance (see Fig.~\ref{fig:lc}d-f). Since DPFs encode additional prior knowledge compared to LSTMs, we might expect them to have higher bias and lower variance. Therefore, DPF's relative error should be lowest with small amounts of data and highest with large amounts of data (the green curves in Fig.~\ref{fig:lc}d-f should go up steadily from left to right until they cross the blue lines). Surprisingly, these curves show a different trend: DPFs relative performance to LSTMs improves with more data and converges to about $\frac{1}{10}$ to $\frac{1}{3}$. There could be a slight upwards trend in the end, but on a logarithmic data axis it would take a tremendous amount of data to close the gap. This result suggests that the priors from the Bayes filter algorithm reduce variance without adding bias---\emph{that these algorithmic priors capture some true structure about the problem}, which data does not help to improve upon.

%----------------------------
\subsubsection{\bf Algorithmic priors lead to policy invariance}
%----------------------------

To be useful for different tasks, localization must be policy-invariant. At the same time, the robot must follow some policy to gather training data, which will inevitably affect the data distribution, add unwanted correlations between states and actions, etc.

We investigated how much the different methods overfit to these correlations by changing the policy between training and test, using two policies A and B. Policy A refers to the heuristic exploration policy that we used for all experiments above (see Sec.~\ref{subsec:problem_setting}). Policy B uses the true pose of the robot, randomly generates a goal cell in the maze, computes the shortest path to the goal, and follows this path from cell to cell using a simple controller mixed with 10\% random actions.

The results in Fig.~\ref{fig:policies} show that all methods have low error rates when tested on their training policy (although DPFs improve over LSTMs even more on policy B). But when we use different policies for training and test, LSTM's error rate jumps to over 80\%, while DPF (ind+e2e) still works in most cases (5\% and 26\% error rate).

The LSTM baseline is not able to generalize to new policies because it does not discriminate between actions and observations and fits to any information that improves state estimation. If the training data includes correlations between states and actions (e.g. because the robot moves faster in a long hallway than in a small room), then the LSTM learns this correlation. Put differently, the LSTM learns to infer the state from the action chosen by the policy. The problem is that this inference fails if the policy changes. The algorithmic priors in DPFs prevent them from overfitting to such correlations because DPFs cannot directly infer states from actions.

DPFs generalize better from A to B than from B to A. Since generalization from B to A is equally difficult for DPFs with individually learned models, the error increase cannot come from overfitting to correlations in the data through end-to-end learning but is most likely because the states visited by policy A cover those visited by policy B but not vice versa.

The alternative approach to encoding policy invariance as a prior is to learn it by adding this variance to the data. Our results show that if we train on combined training data from both policies (A+B), all methods perform well in tests with either policy. This approach in the spirit of domain randomization and data augmentation helps DPFs because it covers the union of the visited states and (additionally) helps LSTM by including state-action correlations from both policies. But to make the LSTM localization truly policy invariant such that it would work with any new policy C, the training data has to cover the space of all policies in an unbiased way, which is difficult for any interesting problem. 

%Therefore, a combination of both approaches is probably most useful.

%To be most data-efficient and invariant to the policy, we should combine algorithmic priors and domain randomization.

\begin{figure}[t]
\centering
\begin{minipage}{.53\columnwidth}  
\subfloat[Visual input (image and difference image) at time steps 100, 200, and 300 (indicated in (b) by black circles)]{\includegraphics[width=\columnwidth]{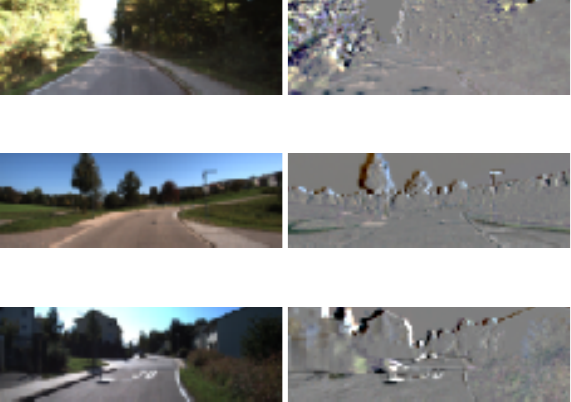}} % 
\vspace{0.0cm}
\end{minipage}
\hfill
\begin{minipage}{.45\columnwidth}  
\subfloat[Trajectory 9; starts at (0,0)]{\includegraphics[width=\columnwidth]{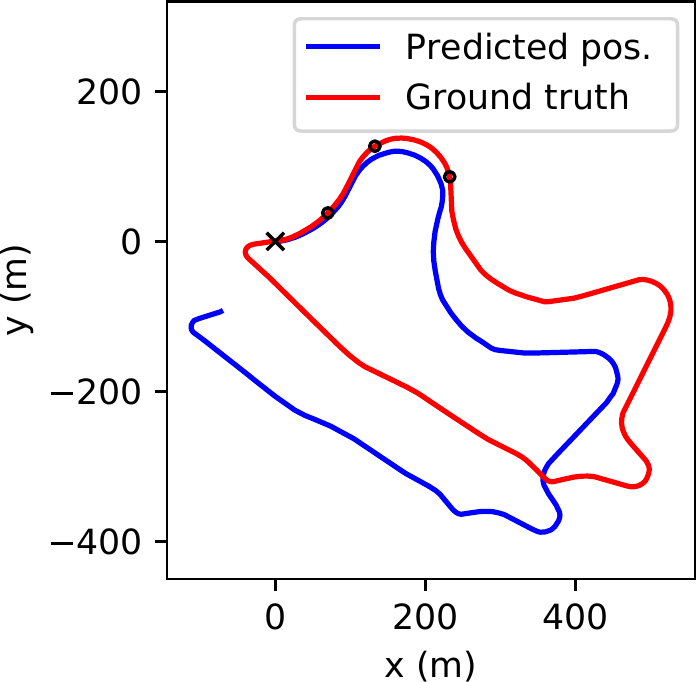}} % 
\end{minipage}
%\hfill
\caption{{\bf Visual odometry with DPFs.} Example test trajectory}
\label{fig:kitti}
\end{figure}

%-------------------------------------------------------------------------------
\subsection{Visual Odometry Task}
%-------------------------------------------------------------------------------

To validate our simulation results on real data, we applied DPFs on the KITTI visual odometry data set, which consists of data from eleven trajectories of a real car driving in an urban area for a total of 40 minutes. The data set includes RGB stereo camera images as well as the ground truth position and orientation of the car in an interval of $\sim$0.1 seconds. The challenge of this task is to generalize in a way that works across highly diverse observations because the method is tested on roads that are never seen during training. Since the roads are different in each trajectory, it is not possible to extract global information about the car's position from the images. Instead, we need to estimate the car's translational and angular velocity from the stream of images and integrate this information over time to track the car's position and orientation. 

We tackle this problem with a DPF in a five dimensional state space, which consists of the position, orientation, forward velocity and angular velocity. DPFs learn to perform visual odometry from a known initial state using a simple first-order \emph{dynamics model}~$g$ and a learnable action sampler $f_{\boldsymbol\theta}$. Since there is no information about the action of the driver, the action sampler produces zero mean motion noise on the velocity dimensions, which is then evaluated with the measurement model. For a fair comparison, we used the same network architecture for the observation encoder $h_{\boldsymbol\theta}$ as in the backprop Kalman filter paper \citep{haarnoja_backprop_2016}, which takes as input the current image and the difference image to the last frame (see Fig.~\ref{fig:kitti}). Our observation likelihood estimator $l_{\boldsymbol\theta}$ weights particles based on their velocity dimensions and the encoding $h_{\boldsymbol\theta}(\boldsymbol o_t)$. Since, the initial state is known, we do not use a particle proposer. We train the DPF individually and end-to-end, using only the velocity dimensions for maximum likelihood estimation.

We evaluated the performance following the same procedure as in the BKF paper. We used eleven-fold cross validation where we picked one trajectory for testing and used all others for training with subsequences of length 50. We evaluated the trained model on the test trajectory by computing the average error over all subsequences of 100 time steps and all subsequences of 100, 200, 400, and 800 time steps.

Table~\ref{table:performance_kitti} compares our results to those published for BKFs~\citep{haarnoja_backprop_2016}. DPFs outperform BKFs, in particular for short sequences where they reduce the error by $\sim$30\%. Any improvement over BKFs in the this task is surprising because Gaussian beliefs seem sufficient to capture uncertainty in this task. The improvement could come from the ability of particles to represent long tailed probability distributions. These results demonstrate that DPFs generalize to different tasks and can be successfully applied to real data.

\begin{table}[t]
%  \centering
  %\renewcommand{\arraystretch}{1.0}
  %\newcommand{\cell}[1]{\framebox[1.5cm][l]{#1}}
  \newcommand{\cell}[1]{\makebox[1.2cm][c]{#1}}
  \newcommand{\cellr}[1]{\makebox[1.5cm][r]{#1}}
  \newcommand{\celll}[1]{\makebox[1.5cm][l]{#1}}
  \newcommand{\titlecell}[1]{\makebox[1cm][l]{#1}}
  \caption{\bf KITTI visual odometry results}
  %For fixed samples, the best performance obtained is shown. For fixed return, the least number of samples required to obtain the desired performance is shown.%
       
  \begin{tabu} to \linewidth {@{} X[1.2, l] X[c] X[c] @{}}
    \toprule
       &
      \cell{Test 100} &
      \cell{Test 100/200/400/800} \\
    \midrule
    \midrule
    \titlecell{Translational error (m/m)} \\
      \quad BKF* & \celll{0.2062} & \celll{0.1804} \\
      \quad DPF (ind) & \celll{0.1901 $\pm$ 0.0229} & \celll{0.2246 $\pm$ 0.0371} \\
            \quad DPF (e2e) & \celll{\bf{0.1467 $\pm$ 0.0149}} & \celll{0.1748 $\pm$ 0.0468} \\
      \quad DPF (ind+e2e) & \celll{0.1559 $\pm$ 0.0280} & \celll{\bf{0.1666 $\pm$ 0.0379}} \\

    \midrule
    \titlecell{Rotational error (deg/m)} \\
      \quad BKF* & \celll{0.0801} & \celll{0.0556} \\
      \quad DPF (ind) & \celll{0.1074 $\pm$ 0.0199} & \celll{0.0806 $\pm$ 0.0153} \\
            \quad DPF (e2e) & \celll{0.0645 $\pm$ 0.0086} & \celll{0.0524 $\pm$ 0.0068} \\
      \quad DPF (ind+e2e) & \celll{\bf{0.0499 $\pm$ 0.0082}} & \celll{\bf{0.0409 $\pm$ 0.0060}} \\
    \midrule

  \titlecell{  Means $\pm$ standard errors; * results from \cite{haarnoja_backprop_2016}} \\
  \end{tabu}

  \label{table:performance_kitti}
\end{table}

%===============================================================================
\section{Conclusion}
\label{sec:discussion}
%===============================================================================

We introduced differentiable particle filters to demonstrate the advantages of combining end-to-end learning with algorithmic priors. End-to-end learning optimizes models for performance while algorithmic priors enable explainability and regularize learning, which improves data-efficiency and generalization.  The use of algorithms as algorithmic priors will help to realize the potential of deep learning in robotics. The components of the DPF implementation, such as sample generation and density estimation, will be useful for producing differentiable versions of other sampling-based algorithms.

\section*{Acknowledgments}

%\acknowledgments{
We gratefully acknowledge financial support by the German Research Foundation (DFG, project number 329426068).
%}

%We proposed to combine algorithmic priors and end-to-end learning. We demonstrated the feasibility and the advantages of this idea in the context of state estimation in robotics. Algorithmic priors lead to data-efficient learning, as knowledge about the problem structure encoded in the algorithm is provided explicitly and does not have to be extracted from data. The ability to learn from data enables the use of algorithms when task-specifics are unknown. The tight combination of both improves performance as the models are optimized for use in the algorithm. We view our results as a proof of concept and are convinced that the combination of algorithms and machine learning will help solve novel problems, while balancing data efficiency and generality.

%===============================================================================

% The archival track will consist of papers up to 8 pages in length (plus up to 2 additional pages of references).

\clearpage
% The acknowledgments are automatically included only in the final version of the paper.
%\acknowledgments{If a paper is accepted, the final camera-ready version will (and probably should) include acknowledgments. All acknowledgments go at the end of the paper, including thanks to reviewers who gave useful comments, to colleagues who contributed to the ideas, and to funding agencies and corporate sponsors that provided financial support.}

%===============================================================================

\balance
\interlinepenalty=10000

\bibliography{references}  % .bib

\end{document}